\documentclass{l4dc2027}

\title[Towards Probabilistically Safe Reinforcement Learning]{Towards Probabilistically Safe Reinforcement Learning: Theory and Deployment on Humanoid and Aerial Robots}

\usepackage{lipsum}
\usepackage{amsmath,amssymb,amsfonts}
\usepackage{algorithm,algorithmic}
\usepackage{hyperref}
\hypersetup{hidelinks=true}
\usepackage{textcomp}
\usepackage{graphicx}
\newsavebox{\rowpanel}
\newlength{\rowsum}
\newlength{\rowH}
\newcommand{\addrowpanel}[2][]{%
  \sbox{\rowpanel}{\includegraphics[height=100pt,#1]{#2}}%
  \addtolength{\rowsum}{\wd\rowpanel}}
\usepackage{verbatim}
\usepackage{esvect}
\usepackage{footnote}
\usepackage{siunitx}
\usepackage{color}
\usepackage{url}
\usepackage[makeroom]{cancel}
\usepackage{todonotes}
\usepackage{subcaption}
\usepackage{bigints}
\usepackage{booktabs}
\usepackage{arydshln}  
\usepackage{array}
\usepackage{multirow}
\usepackage{enumitem}

\newcommand{\real}{\mathbb{R}}

\newcommand{\Cc}{{\mathcal{C}}}

\newcommand{\Rc}{{\mathcal{R}}}

\newcommand{\Sc}{{\mathcal{S}}}
\newcommand{\Ac}{{\mathcal{A}}}

\newcommand{\bs}{{s}}

\newcommand{\by}{{y}}
\newcommand{\ba}{{a}}
\newcommand{\bz}{{z}}
\newcommand{\be}{{e}}

\newcommand{\bp}{{p}}
\newcommand{\bq}{{q}}

\newcommand{\bc}{{c}}

\newcommand{\btheta}{{\theta}}
\newcommand{\bxi}{{\xi}}

\newcommand{\R}{\mathbb{R}}
\newcommand{\E}{\mathbb{E}}

\newcommand{\gh}[1]{\widehat{\nabla V}_{#1}}
\newcommand{\Vh}{\widehat V_1}
\newcommand{\pos}[1]{[#1]_+}

\newcommand\xqed[1]{%
  \leavevmode\unskip\penalty9999 \hbox{}\nobreak\hfill
  \quad\hbox{#1}}
\newcommand\demo{\xqed{$\bullet$}}

\DeclareMathOperator*{\argmin}{arg\,min}

\newcommand{\longthmtitle}[1]{\mbox{}\emph{(#1):}}
\newcommand{\setdef}[2]{\{#1 : #2\}}

\newcommand{\norm}[1]{\left\lVert#1\right\rVert}

\allowdisplaybreaks

\author{%
 \Name{Pol Mestres} \Email{mestres@caltech.edu} \\
 \Name{Varun Gabbita} \Email{vgabbita@caltech.edu} \\
 \Name{Jason H.K. Siu}
 \Email{jasonsiu@caltech.edu} \\
 \Name{Sergio A. Esteban}
 \Email{sesteban@caltech.edu} \\
 \Name{Hampus Serneke} \Email{hserneke@caltech.edu} \\
 \Name{Joel W. Burdick}
 \Email{jburdick@caltech.edu} \\
 \Name{Aaron D. Ames}
 \Email{ames@caltech.edu} \\
 \addr Department of Mechanical and Civil Engineering, Caltech, Pasadena, CA, USA
}

\begin{document}


\maketitle

\begin{abstract}
Reinforcement learning (RL) enables robots to acquire complex skills, but standard RL formulations do not guarantee that learned policies satisfy safety constraints, which limits their use in safety-critical settings. Existing safe RL methods with provable guarantees typically support only restricted classes of constraints and require unbiased estimators of the objective and constraint functions, which prevents their direct integration with state-of-the-art algorithms such as proximal policy optimization (PPO). In this paper, we extend the recently introduced robust safe gradient flow-based reinforcement learning (RSGF-RL) framework to overcome these limitations. Our formulation accommodates a general class of trajectory-level safety constraints, including chance constraints such as collision avoidance with a prescribed probability, and our analysis allows for biased value and gradient estimators.
We show that, with high probability, the constraint violation throughout training is bounded by the estimator bias plus a term that vanishes as the number of rollouts grows, and decays to this level if the initial policy is unsafe. We also establish the convergence properties of the iterates to approximate KKT points of the original safe RL problem. Finally, we show that RSGF-RL implemented with PPO-style estimates enables zero-shot sim-to-real deployment of whole-body control policies on a Unitree G1 humanoid and collision-free waypoint navigation on a Crazyflie drone.
\end{abstract}


\section{Introduction}

Reinforcement learning (RL) has emerged as a powerful tool for enabling robots to acquire a wide range of complex skills~\citep{JH-JL-AD-DB-VT-VK-MH:19,OA-MA-BB:20,EK-LB-AL-MM-VK-DS:23}.
However, standard RL algorithms primarily optimize task performance, without guaranteeing that the learned policies satisfy safety constraints, either during training or after deployment.
This lack of formal safety guarantees remains a major obstacle to the adoption of RL in safety-critical applications.

Motivated by this challenge, a growing body of work on \emph{safe RL} seeks to optimize policy performance subject to prescribed safety specifications~\citep{LB-MG-AWH-ZY-SZ-JP-APS:22,JG-FF:15,SG-LY-YD-GC-FW-JW-AK:24}.
A common approach formulates safe RL as a constrained policy optimization problem, where the constraint requires the expected cumulative value of a safety-related cost to remain below a prescribed threshold~\citep{JA-DH-AT-PA:17,YC-MG-LJ-MP:17,SP-LC-MCF-AR:19}.
Several algorithms have been proposed to solve such problems, including primal-dual methods~\citep{SP-MCF-LFOC-AR:23,DD-KZ-TB-MJ:20}, constrained policy optimization (CPO)~\citep{JA-DH-AT-PA:17}, projection-based constrained policy optimization (PCPO)~\citep{TY-JR-KN-PJR:20}, and state augmentation~\citep{AS-AIC-TJ-ZW-DHM-JW-HA:22}.
Despite their empirical success, these methods generally do not guarantee constraint satisfaction throughout the learning process, and their convergence guarantees rely on restrictive assumptions.

An alternative approach enforces safety at the level of individual control actions, typically by combining an unconstrained RL policy with a safety filter based on control barrier functions (CBFs)~\citep{FPB-LB-APS:25,MA-RB-RE-BK-SN-UT:18,RC-GO-RMM-JWB:19,LY-BW-MdS-ADA:25,KH-HH-JFF:24}.
While such methods can provide guarantees on the safety of the resulting closed-loop system, they do not directly address the problem of optimizing a policy subject to safety constraints.

In this paper, we adopt the constrained policy optimization perspective, building on the recently introduced \emph{robust safe gradient flow-based reinforcement learning} (RSGF-RL) framework~\citep{PM-AM-JC:25-l4dc,PM-AM-JC:25-arxiv}.
RSGF-RL provides convergence guarantees to critical points of the constrained RL problem while maintaining constraint satisfaction throughout the learning process with high probability.
However, several limitations restrict its
integration with state-of-the-art RL algorithms and modern robotic systems:

\begin{enumerate}
    \item \textbf{Limited class of safety constraints.}
    The existing framework considers constraints that require that the expected value of \textit{cumulative safety costs} remains below a given threshold.
    However, many safety specifications arising in robotics are naturally expressed as \emph{probabilistic constraints}, such as requiring that the probability of colliding with an obstacle during an episode remains below a prescribed threshold, which are not equivalent to the expected cumulative cost formulations.
    \item \textbf{Restrictive assumptions.}
    The theoretical guarantees of RSGF-RL as provided in~\citep{PM-AM-JC:25-l4dc,PM-AM-JC:25-arxiv} require \textit{unbiased estimators} of the objective and constraint functions and their gradients.
    In contrast, modern RL algorithms such as Proximal Policy Optimization (PPO)~\citep{JS-FW-PD-AR-OK:17}, Soft Actor-Critic (SAC)~\citep{TH-AZ-PA-SL:18}, and DreamerV3~\citep{DH-JP-JB-TL:25} employ estimation procedures that can introduce bias.
    This limits the integration of RSGF-RL with state-of-the-art policy optimization methods. Additionally, the conditions under which~\citep{PM-AM-JC:25-arxiv} establishes convergence of RSGF-RL (such as the requirement that the estimated RSGF-RL update law converges to the true one with probability one) are quite restrictive.
    \item \textbf{Lack of hardware validation.}
    Previous evaluations of RSGF-RL have been limited to relatively simple simulation environments, leaving open whether its theoretical guarantees can be combined with the computational demands and practical challenges of RL policies for real robotic systems.
\end{enumerate}

In this paper, we develop an extension of RSGF-RL that addresses these shortcomings, and show that the resulting algorithm can be used to train RL policies for a variety of different tasks, with zero-shot sim-to-real transfer.
Our contributions are as follows.

\begin{enumerate}
    \item First, we extend RSGF-RL to a general class of trajectory-level constraints, which includes cumulative
    cost and probabilistic constraints. Similarly to the Policy Gradient Theorem~\citep[Chapter 13]{RSS-AGB:18}, we show that gradient estimates for this class of constraints follow from the classical likelihood-ratio method in~\citep{PWG:90};
    \item Second, we show that, with high probability, the constraint violation of the RSGF-RL iterates is bounded by the estimator bias plus a sampling-error term that vanishes as the number of rollouts $N$ used for value and gradient estimation grows.
    If the initial policy is safe, this bound holds throughout training; if it is unsafe, the violation decays to this level.
    We also show that, almost surely, the RSGF-RL iterates have a limit point that is an approximate KKT point of the constrained RL problem, with the approximation level again given by the estimator bias plus a term that vanishes as $N\to\infty$.
    \item Third, we show that RSGF-RL implemented with (biased) PPO gradient estimates enables zero-shot sim-to-real deployment of whole-body motor skills on a Unitree G1 humanoid robot and collision-free waypoint navigation on a Crazyflie drone. Some assumptions of our theoretical results are not met in the practical implementation of RSGF-RL used in these experiments. We discuss this theory-to-practice gap in Section~\ref{sec:theory-practice-gap}.
\end{enumerate}


\section{Background}

\subsection{Safe Reinforcement Learning}

Consider a Markov Decision Process $(\mathcal{S},\mathcal{A},P)$ with unknown transition function $P$. Let $T\in\mathbb{N}$ be a finite time horizon, and let
$g_0,g_1:\mathcal{S}^{T+2}\times\mathcal{A}^{T+1}\to\real$
be real-valued functions.
We also consider a distribution $\eta$ over initial states and a parametric class of policies indexed by $\btheta\in\real^d$, with $\pi_{\btheta}$ denoting the policy associated with parameter $\btheta$.
Let $\bs=[s_0,\hdots,s_{T+1}]$ and $\ba=[a_0,\hdots,a_T]$. We consider the constrained RL problem
\begin{align}\label{eq:safe-RL-problem}
    \min_{\btheta\in\real^d} \quad & V_0(\btheta) = \mathbb{E}[g_0(\bs,\ba)] \qquad
    \text{s.t.} \quad V_1(\btheta) = \mathbb{E}[g_1(\bs,\ba)] \leq 0.
\end{align}
The expectations in~\eqref{eq:safe-RL-problem} are taken with respect to the probability measure over trajectories induced by the initial-state distribution $\eta$, the policy $\pi_{\btheta}$, and the transition function $P$. Unless stated otherwise, all expectations are taken with respect to this measure.

The function $g_0$ encodes the objective to be minimized over the horizon $T$, whereas $g_1$ encodes a safety constraint over the same horizon.
For simplicity, we consider~\eqref{eq:safe-RL-problem} with a single safety constraint and same time horizon for both the objective and constraint, but the results developed in the paper extend naturally to multiple constraints and different time horizons.

We next present different instances of~\eqref{eq:safe-RL-problem} that illustrate the generality of this formulation and connect it with safe RL problems previously studied in the literature.

\begin{example}\longthmtitle{Cumulative constraints}\label{ex:cumulative-constraints}
    Let $\gamma\in(0,1]$ be a discount factor,
    $R_0:\Sc\times\Ac\times\Sc\to\real$ a reward function, and
    $R_1:\Sc\times\Ac\times\Sc\to\real$ a cost function.
    Consider $g_j(\bs,\ba) = (-1)^{j+1}\sum_{t=0}^T \gamma^t R_j(s_t,a_t,s_{t+1})$.
    Then~\eqref{eq:safe-RL-problem} becomes
    the standard constrained RL formulation considered, e.g., in~\citep{EA:99,JA-DH-AT-PA:17,YC-MG-LJ-MP:17,SP-LC-MCF-AR:19,PM-AM-JC:25-l4dc}.
    The objective is to maximize the expected discounted cumulative reward while constraining the expected discounted cumulative cost.
    \demo
\end{example}

\begin{example}\longthmtitle{Probabilistic constraints}\label{ex:probabilistic-constraints}
    Let $\bar{\Cc}\subset\Sc$ be a safe set and let $\delta\in(0,1)$.
    Taking $g_0$ as in Example~\ref{ex:cumulative-constraints} and defining
    $g_1(\bs,\ba) = -\mathbb{I}\Big(
    \bigcap_{t=0}^T\{s_t\in\bar{\Cc}\} \Big) +1-\delta$, where $\mathbb{I}(A)$ denotes the indicator function of an event $A$,
    problem~\eqref{eq:safe-RL-problem} becomes
    \begin{align}\label{eq:probabilistic-constraint-problem}
        &\max_{\btheta\in\real^d}
        \ \mathbb{E}\Big[\sum_{t=0}^T
        \gamma^t R_0(s_t,a_t,s_{t+1})\Big], \qquad
        \text{s.t.} \quad
        \mathbb{P}\Big(
        \bigcap_{t=0}^T\{s_t\in\bar{\Cc}\}
        \Big)
        \geq 1-\delta.
    \end{align}
    This formulation requires the state trajectory to remain in the safe set $\bar{\Cc}$ over the entire horizon with probability at least $1-\delta$, and is considered, for example, in~\citep{WC-DS-SP:24,SP-MCF-LFOC-AR:23}.
    As shown in Appendix~\ref{sec:appendix-connection-cumulative-probabilistic}, an appropriate choice of $R_1$ in Example~\ref{ex:cumulative-constraints} can also imply the probabilistic constraint in~\eqref{eq:probabilistic-constraint-problem}, but the two formulations are not equivalent.
    \demo
\end{example}


Beyond examples~\ref{ex:cumulative-constraints}-\ref{ex:probabilistic-constraints}, problem~\eqref{eq:safe-RL-problem} also covers the percentile risk constraints considered in~\citep{YC-MG-LJ-MP:17,CY-XZ-DY-JZ:21} and the conditional chance constraints in~\citep{XS-SJ-AW-KH-SG:24}.
Hence,~\eqref{eq:safe-RL-problem} captures several common notions of safety in RL.

\subsection{Robust Safe Gradient Flow}

We next introduce the Robust Safe Gradient Flow (RSGF)~\citep{PM-AM-JC:25-arxiv}, a constrained optimization method that will serve as the basis for solving~\eqref{eq:safe-RL-problem}.
Let $V_0,V_1:\real^d\to\real$ be continuously differentiable functions and consider the constrained optimization problem
\begin{align}\label{eq:V0-V1-optimization-problem}
    \min_{\btheta\in\real^d} V_0(\btheta) \qquad
    \text{s.t.}\quad V_1(\btheta)\leq 0.
\end{align}
Let $\Cc=\setdef{\btheta\in\real^d}{V_1(\btheta)\leq 0}$
denote its feasible set.
Given $\alpha,\beta>0$, define the RSGF vector field
$\Rc:\real^d\to\real^d$ by

\begin{equation}\label{eq:R-definition}
\begin{aligned}
    \Rc(\btheta)
    =
    \argmin_{\bxi\in\real^d}
    \frac{1}{2}
    \norm{\bxi+\nabla V_0(\btheta)}^2
    \quad
    \text{s.t.}\quad
    \alpha V_1(\btheta)
    +
    \nabla V_1(\btheta)^\top\bxi
    +
    \frac{\beta}{2}\norm{\bxi}^2
    \leq 0.
\end{aligned}
\end{equation}
Thus, $\Rc(\btheta)$ selects a direction that remains as close as possible to the negative gradient $-\nabla V_0(\btheta)$ while satisfying a constraint designed to preserve feasibility.
Given a sequence of positive stepsizes $\{h_t\}_{t\in\mathbb{N}}$, we consider the discrete-time RSGF iteration
\begin{align}\label{eq:discrete-time-rsgf}
    \btheta_{t+1}
    =
    \btheta_t
    +
    h_t\Rc(\btheta_t).
\end{align}
As shown in~\citep[Lemma 1]{PM-AM-JC:25-l4dc} and~\citep[Lemma IV.2, Prop. IV.3]{PM-AM-JC:25-arxiv} (consolidated in Proposition~\ref{prop:rsgf-properties} in Appendix~\ref{sec:details-proof-Proposition-RSGF-properties} for completeness), under a constraint qualification and suitable Lipschitz and stepsize conditions, the 
limit points of~\eqref{eq:discrete-time-rsgf} are KKT points of~\eqref{eq:V0-V1-optimization-problem}.
Besides ensuring convergence to KKT points of~\eqref{eq:V0-V1-optimization-problem}, this result highlights the \textit{anytime nature of RSGF}: provided that the initial iterate is feasible and the stepsizes satisfy the stated conditions, every subsequent iterate remains feasible.


\section{Safe Reinforcement Learning with the Robust Safe Gradient Flow}

Although the RSGF iterates~\eqref{eq:discrete-time-rsgf} have KKT points of the constrained optimization problem~\eqref{eq:safe-RL-problem} as limit points, they cannot be applied directly in the model-free RL setting. In particular, because 
the transition kernel $P$ of the MDP is unknown, the value functions $V_0,V_1$ and their gradients $\nabla V_0,\nabla V_1$ cannot in general be evaluated exactly. In this section, we therefore construct estimates of these quantities from policy rollouts.

The following result provides an expression for the gradient of the general functions $V_0$ and $V_1$ introduced in~\eqref{eq:safe-RL-problem}, and will serve as the basis for estimating such gradients from sampled trajectories.
This result is a direct consequence of the likelihood-ratio gradient identity~\citep{PWG:90}, and its proof is deferred to Appendix~\ref{sec:appendix-proof-policy-gradient-general}, where we also discuss the special cases it recovers (including the standard Policy Gradient Theorem~\citep[Chapter 13]{RSS-AGB:18}).


\begin{proposition}\longthmtitle{Policy gradient for general objective and constraints}\label{prop:policy-gradient-general}
    Suppose that, for every $a\in\Ac$ and $s\in\Sc$, the mapping
    $\btheta \mapsto \log\pi_{\btheta}(a|s)$ is continuously differentiable, the functions $g_0, g_1$ are uniformly bounded, and
    for every $\btheta\in\real^d$, there exists a neighborhood $U_{\btheta}$ and a constant $M_{\btheta}>0$ where 
    \begin{align}\label{eq:uniform-boundedness-logpitheta}
        \sup\limits_{ \btheta^\prime \in U_{\btheta} } \sup\limits_{s\in\Sc, a\in\Ac} \norm{\nabla_{\btheta^\prime}\log\pi_{\btheta^\prime}(a|s) } \leq M_\btheta.
    \end{align}
    Let $H_t=(s_0,a_0,s_1,a_1,\hdots,a_{t-1},s_t)\in(\Sc\times\Ac)^t\times\Sc$
    denote the trajectory history available when action $a_t$ is selected.
    Further,\footnote{For non-negative integers $n_1,n_2$, we use the notation $n_1:n_2=\{n_1,n_1+1,\hdots,n_2\}$.}, for each $t\in0:T$, let
    $b_{j,t}:(\Sc\times\Ac)^t\times\Sc\times\real^d\to\real$
    be an arbitrary integrable function.
    Then, for $j\in\{0,1\}$, 
    \begin{align}\label{eq:gradient-expression}
        \nabla V_j(\btheta)
        =
        \mathbb{E}\Big[
        \sum_{t=0}^T
        \big(g_j(\bs,\ba)-b_{j,t}(H_t,\btheta)\big)
        \nabla_{\btheta}\log\pi_{\btheta}(a_t|s_t)
        \Big].
    \end{align}
\end{proposition}


Fix $j\in\{0,1\}$. Given $N\in\mathbb{N}$ independent episodes generated under $\pi_\btheta$ with initial distribution $\eta$, empirical averages of $g_j$ and~\eqref{eq:gradient-expression} yield unbiased estimators of $V_j$ and $\nabla V_j$, respectively.
Practical RL algorithms, however, often modify these estimators to reduce variance or improve numerical performance, generally at the cost of bias: PPO-clip~\citep[Section 3]{JS-FW-PD-AR-OK:17} optimizes a clipped surrogate objective, and actor--critic methods~\citep{VRK-JNT:03} build lower-variance gradient estimates from learned value functions.
We therefore consider general estimators
\begin{align}\label{eq:general-estimates}
    \widehat{V}_j(\btheta) = \frac{1}{N}\sum_{n=1}^N G_{j,n}^{\btheta},
    \qquad
    \widehat{\nabla V}_j(\btheta) = \frac{1}{N}\sum_{n=1}^N \tilde{G}_{j,n}^{\btheta},
\end{align}
where $G_{j,n}^{\btheta}\in\real$ and $\tilde{G}_{j,n}^{\btheta}\in\real^d$ satisfy, for all $n\in 1:N$, $i\in 1:d$, and some constants $\zeta_j,\tilde\zeta_j \geq 0$ and $B_j,\tilde B_j$ (independent of $\theta$),
\begin{align}
    \big|\mathbb{E}[G_{j,n}^{\btheta}]-V_j(\btheta)\big| &\leq \zeta_j,
    &
    \big|\mathbb{E}[(\tilde{G}_{j,n}^{\btheta})_i]-(\nabla V_j(\btheta))_i\big| &\leq \tilde\zeta_j,
    \label{eq:bounded-bias}
    \\
    |G_{j,n}^{\btheta}| &\leq B_j,
    &
    |(\tilde{G}_{j,n}^{\btheta})_i| &\leq \tilde B_j \quad \text{a.s.}
    \label{eq:boundedness-estimates}
\end{align}
Assumption~\eqref{eq:boundedness-estimates} is standard in RL~\citep{KZ-AK-HZ-TB:20,QB-WUM-VA:24,PM-AM-JC:25-l4dc,PM-AM-JC:25-arxiv}; for the estimators induced by~\eqref{eq:gradient-expression}, it holds whenever $\nabla_{\btheta}\log\pi_{\btheta}$, $\{g_j\}$, and $\{b_{j,t}\}$ are uniformly bounded over $(\bs,\ba)$. Bounds of the form~\eqref{eq:bounded-bias} are available for PPO~\citep[Theorem 4.2]{LD-DS-MM-SK-BW-TA-SW:26} and actor--critic algorithms~\citep[Lemma 4.10]{SM-DK:24}.
Using~\eqref{eq:general-estimates}, we define the empirical RSGF direction
\begin{equation}\label{eq:R-hat-definition}
\begin{aligned}
    \hat{\Rc}(\btheta)
    =
    \argmin_{\bxi\in\real^d}
    \frac{1}{2}
    \norm{\bxi+\widehat{\nabla V_0}(\btheta)}^2
    \quad
    \text{s.t.}\quad
    \alpha\widehat{V}_1(\btheta)
    +
    \widehat{\nabla V}_1(\btheta)^\top\bxi
    +
    \frac{\beta}{2}\norm{\bxi}^2
    \leq0,
\end{aligned}
\end{equation}
and the stochastic policy update
\begin{align}\label{eq:discrete-time-estimated-RSGF}
    \btheta_{t+1}
    =
    \btheta_t
    +
    h_t\hat{\Rc}(\btheta_t),
\end{align}
where for every $\theta_t$, $\hat{\Rc}(\theta_t)$ is computed through~\eqref{eq:R-hat-definition} (with $\theta=\theta_t$) and estimates obtained through $N$ independent rollouts as in~\eqref{eq:general-estimates}, and $\{ h_t \}$ is a deterministic sequence of stepsizes.
We note that $\hat{\Rc}(\theta)$ (and $\Rc(\theta)$) can be computed in closed-form (cf.~\citealp[Lemma 3]{PM-AM-JC:25-l4dc}).
We refer to the algorithm that updates policy parameters according to~\eqref{eq:discrete-time-estimated-RSGF} as RSGF-RL.
If problem~\eqref{eq:R-hat-definition} is infeasible at $\theta=\theta_t$, we set
$\hat{\Rc}(\theta_t):=-\gh1(\theta_t)/\beta$.
We next investigate the extent to which RSGF-RL retains the safety and convergence properties of the nominal RSGF iterates~\eqref{eq:discrete-time-rsgf}. 
In the following results, we let $\mathcal{F}_t$ be the filtration obtained from all stochasticity up to iteration $t$ (but not including the rollouts from $\pi_{\theta_t}$).
We begin by establishing a bound on the number of samples required to achieve a probabilistic safety guarantee.

\begin{proposition}\longthmtitle{Probabilistic safety}\label{prop:safety-RSGF-RL}
    Consider a sequence $\{\btheta_t\}_{t\in\mathbb{N}}$ generated by~\eqref{eq:discrete-time-estimated-RSGF}, and let $K\in\mathbb{N}$.
    Suppose that $\nabla V_1$ is $L_1$-Lipschitz, that the stepsizes 
    satisfy $0<h_t\le h_{\max}<\min\{1/\alpha,\beta/L_1\}$, and that, almost surely,~\eqref{eq:R-hat-definition} is feasible at $\btheta=\btheta_t$ for every $t\in 0:K-1$.
    Fix $\epsilon>0$ and $\bar\delta\in(0,1)$, and let
    \begin{equation}\label{eq:N-bound-safety}
    N \ge \max\Big\{\tfrac{2B_1^2\log(4/\bar\delta)}{\epsilon^2},\;
                   \tfrac{2d\tilde B_1^2\log(4d/\bar\delta)}{\epsilon^2}\Big\}.
    \end{equation}
    Define $\Pi_t := \prod_{k=0}^{t-1}(1-\alpha h_k)$ and
    \begin{equation}\label{eq:cbar}
    \bar c := \zeta_1+\epsilon+\frac{(\sqrt{d}\,\tilde\zeta_1+\epsilon)^2}{2\alpha(\beta-L_1h_{\max})}.
    \end{equation}
    Then, with probability at least $1-K\bar\delta$,
    \begin{equation}\label{eq:multistep}
    V_1(\btheta_t) \le \Pi_t\, V_1(\btheta_0) + (1-\Pi_t)\,\bar c
    \qquad\text{for all } t\in 0:K.
    \end{equation}
    In particular, with probability at least $1-K\bar\delta$:
    (i) if $V_1(\btheta_0)\le 0$, then $V_1(\btheta_t)\le\bar c$ for all $t\in 0:K$;
    (ii) if $V_1(\btheta_0)>0$, then $V_1(\btheta_t)\le \Pi_t V_1(\btheta_0)+\bar c$ for all $t\in 0:K$. 
\end{proposition}

The proof of Proposition~\ref{prop:safety-RSGF-RL} follows an argument similar to that of~\citep[Proposition 5]{PM-AM-JC:25-l4dc} and is included for completeness in Appendix~\ref{sec:proof-proposition-safety-RSGF-RL}. 
Proposition~\ref{prop:safety-RSGF-RL} quantifies how estimation error and bias affect the safety of policy updates. 
It shows that with high probability (which degrades linearly with the safety horizon length $K$), (i) RSGF-RL iterates with safe initial policy stay in a neighborhood of the safe set (with the size of the neighborhood determined by $\bar{c}$, which shrinks as the estimator bias decreases and the number of rollouts $N$ increases); (ii) RSGF-RL iterates with unsafe initial policy exponentially converge (e.g. by choosing constant stepsizes) to this neighborhood of the safe set.
Note that if we want to ensure exact constraint satisfaction with high probability, we can simply tighten the constraint in~\eqref{eq:V0-V1-optimization-problem} by a margin $\bar{c}$.


We next study the convergence properties of~\eqref{eq:discrete-time-estimated-RSGF}.
For $\theta\in\R^d$ define
\begin{equation}\label{eq:radius}
  D(\theta) := \frac{\|\nabla V_1(\theta)\|^2}{\beta^2}-\frac{2\alpha V_1(\theta)}{\beta},
  \qquad
  \hat D(\theta) := \frac{\|\gh1(\theta)\|^2}{\beta^2}-\frac{2\alpha \Vh(\theta)}{\beta}.
\end{equation}
As shown in Lemma~\ref{lem:ball}, problem~\eqref{eq:R-definition} (resp.~\eqref{eq:R-hat-definition}) is feasible if and only if $D(\theta)\geq0$ (resp. $\hat D(\theta)\geq0$).

 
 
\begin{proposition}\longthmtitle{Convergence}\label{prop:conv}
Suppose that:
 
\begin{enumerate}[label=(\roman*), ref=(\roman*)]
  \item\label{it:Theta-compact-convex} $\Theta\subset\R^d$ is compact and convex, and $\theta_t\in\Theta$ for all $t\in\mathbb{N}$ almost surely;
  \item\label{it:Lipschitz-gradients} $\nabla V_0,\nabla V_1$ are $L_0$-, $L_1$-Lipschitz on $\Theta$;
  \item\label{it:rho-feasibility-margin} problem~\eqref{eq:R-definition} is feasible for all $\theta\in\Theta$ and $\rho:=\min_{\theta\in\Theta} \sqrt{D(\theta)}>0$;
\end{enumerate}
 
Define the constants $G_j:=\max_{\theta\in\Theta}\|\nabla V_j(\theta)\|$ for $j\in\{0,1\}$,
\begin{gather}
  \bar u := \frac{(G_0+G_1/\beta)^2}{\beta\rho^2},\qquad
  c_1 := \frac1\beta+\frac{\sqrt d\,\tilde B_1+G_1}{\beta^2\rho},\qquad
  c_V := \frac{2\alpha}{\beta\rho},\label{eq:consts}\\
  \varepsilon_N^2 := 3\Big[d\big(\tilde\zeta_0^2+\tfrac{\tilde B_0^2}{N}\big)
     + c_1^2\, d\big(\tilde\zeta_1^2+\tfrac{\tilde B_1^2}{N}\big)
     + c_V^2\big(\zeta_1^2+\tfrac{B_1^2}{N}\big)\Big].\label{eq:epsN}
\end{gather}
Fix any $\kappa\ge\bar u$, and let
$\Phi_\kappa(\theta):=V_0(\theta)+\kappa\pos{V_1(\theta)}$,
$G_\Phi:=G_0+\kappa G_1$, $L_\Phi:=L_0+\kappa L_1$,
$\Delta_0:=\Phi_\kappa(\theta_0)-\min_\Theta V_0$, and $H_K:=\sum_{t=0}^{K-1}h_t$ for $K\in\mathbb{N}$.
%
Let $\{\theta_t\}$ be generated by~\eqref{eq:discrete-time-estimated-RSGF} with
$0<h_t\le\bar h:=\min\{1/\alpha,\,1/(2L_\Phi)\}$. Then:
\begin{enumerate}[label=(\alph*), ref=(\alph*)]
\item\label{it:non-asymptotic-bound} \emph{(Non-asymptotic bound)} For every  $K\in\mathbb{N}$, 
\begin{equation}\label{eq:main}
  \frac{1}{H_K}\sum_{t=0}^{K-1}h_t\,\E\|\Rc(\theta_t)\|^2
  \;\le\; \frac{2\Delta_0}{H_K} + 2G_\Phi\,\varepsilon_N
  + \frac{2L_\Phi\varepsilon_N^2}{H_K}\sum_{t=0}^{K-1}h_t^2 .
\end{equation}
In particular, for a constant stepsize $h_t\equiv h\le\bar h$,
\begin{equation}\label{eq:const}
  \min_{0\le t<K}\E\|\Rc(\theta_t)\|^2\;\le\;\frac1K\sum_{t=0}^{K-1}\E\|\Rc(\theta_t)\|^2
  \;\le\; \frac{2\Delta_0}{hK} + 2G_\Phi\,\varepsilon_N+\varepsilon_N^2 .
\end{equation}
\item\label{it:almost-sure-bound} \emph{(Almost-sure asymptotic bound)} If $\sum_t h_t=\infty$, then almost surely
\begin{equation}\label{eq:as}
  \limsup_{K\to\infty}\frac{1}{H_K}\sum_{t=0}^{K-1}h_t\|\Rc(\theta_t)\|^2
  \;\le\; 2G_\Phi\,\varepsilon_N+\varepsilon_N^2 ,
\end{equation}
so in particular $\liminf_{t\to\infty}\|\Rc(\theta_t)\|^2\le 2G_\Phi\varepsilon_N+\varepsilon_N^2$ a.s.
\end{enumerate}
\end{proposition}

The proof of Proposition~\ref{prop:conv} is included in Appendix~\ref{app:proof-proposition-conv}.
Let us comment on the assumptions of Proposition~\ref{prop:conv}.
First,~\ref{it:Theta-compact-convex} can easily be guaranteed by adding an additional deterministic constraint in~\eqref{eq:R-hat-definition} that guarantees that $\Theta$ is forward invariant, as done in~\cite{PM-AM-JC:25-arxiv}.
On the other hand,~\ref{it:Lipschitz-gradients} is standard in the literature, and the Lipschitz constants can even be obtained explicitly as in~\cite[Lemma 3.2]{KZ-AK-HZ-TB:20}.
Finally, since feasibility of problem~\eqref{eq:R-definition} at parameter $\theta$ is equivalent to $D(\theta) \geq 0$, condition~\ref{it:rho-feasibility-margin} requires that problem~\eqref{eq:R-definition} has a strict feasibility margin in the set $\Theta$.
%

Since $\Theta$ is compact and $\Rc$ is continuous, Proposition~\ref{prop:conv}\ref{it:almost-sure-bound} implies that, almost surely, a subsequence of the RSGF-RL iterates~\eqref{eq:discrete-time-estimated-RSGF} converges to a point 
which satisfies the KKT conditions of~\eqref{eq:V0-V1-optimization-problem} up to a tolerance that vanishes with $\varepsilon_N$. Hence, as $\varepsilon_N \to 0$ (equivalently, as the estimator biases approach $0$ and the number of rollouts $N$ increases), such points form an approximate KKT sequence as introduced in~\citep{RA-GH-JMM:11}, whose limit points are KKT points of~\eqref{eq:V0-V1-optimization-problem} (because assumption~\ref{it:rho-feasibility-margin} implies MFCQ).
In particular, if estimates are constructed so that they become asymptotically unbiased, the approximation error only depends on the number of rollouts $N$.

\section{Experiments}\label{sec:results}
We implement RSGF-RL~\eqref{eq:discrete-time-estimated-RSGF} for humanoid whole-body control and drone obstacle avoidance tasks\footnote{Videos of the different experiments are available at the following link: \url{https://drive.google.com/file/d/19_EAcHGyLCidhLtGAe3-6qqPSirQZppJ/view?usp=sharing}.}.
For all tasks we use Gaussian policies with mean parametrized by a multi-layer perceptron (MLP). 
Furthermore, the policy acts on observations rather than on the full state; this fits our framework by augmenting the state with such observations\footnote{In this setting, conditions~\eqref{eq:uniform-boundedness-logpitheta} and~\eqref{eq:boundedness-estimates} do not hold for Gaussian policies with unbounded support, but they hold for Gaussian policies truncated to a compact action set $\Ac$ with standard deviation bounded away from zero on the compact parameter space $\Theta$ and observations bounded on a compact set. Truncating actions and observations on a sufficiently large set retains the same performance in practice.}.
%
We obtain estimates $\widehat{\nabla V}_0$ through PPO-clip, and estimates $\widehat{\nabla V}_1, \widehat{V}_1$ through an actor-critic scheme.
Both provide biased estimates (cf.~\citealp[Theorem 4.2]{LD-DS-MM-SK-BW-TA-SW:26} and~\citealp[Theorem 2]{HK-AK-AR:23} for bias bounds).
%
Since the choice of $g_1$ in Example~\ref{ex:probabilistic-constraints} tends to stall learning, possibly due to its sparse nature, we use $g_0, g_1$ as in Example~\ref{ex:cumulative-constraints}, with $R_1$ as in~\eqref{eq:R1-practical-choice-probabilistic-safety}.
Rewards, hyperparameters, and details on how the estimates are computed are reported in Appendices~\ref{sec:experiments-reward-definitions} and~\ref{sec:training-parameters}. 

\subsection{Humanoid whole-body control}

\begin{figure}[t]
    \renewcommand*{\subfigurelabel}[1]{{\scriptsize(\emph{#1})}}%
    \centering
    \setlength{\rowsum}{0pt}%
    \addrowpanel{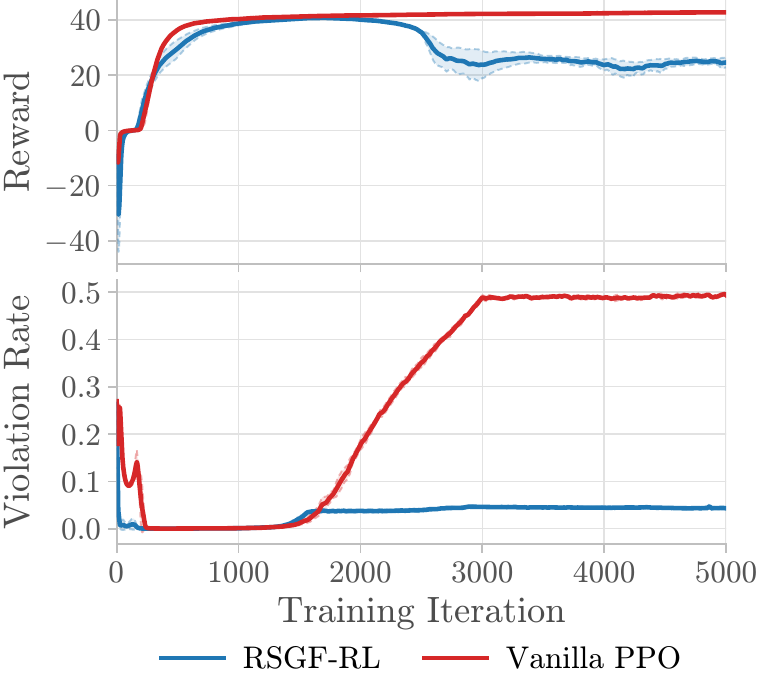}%
    \addrowpanel{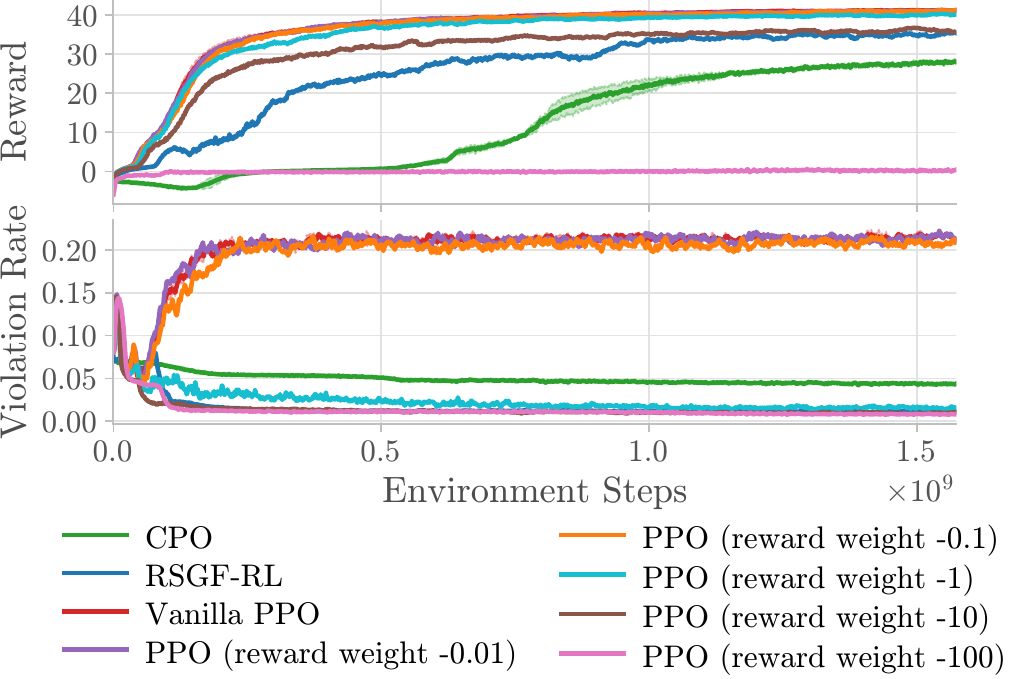}%
    \addrowpanel[trim=19.4bp 0 0 0,clip]{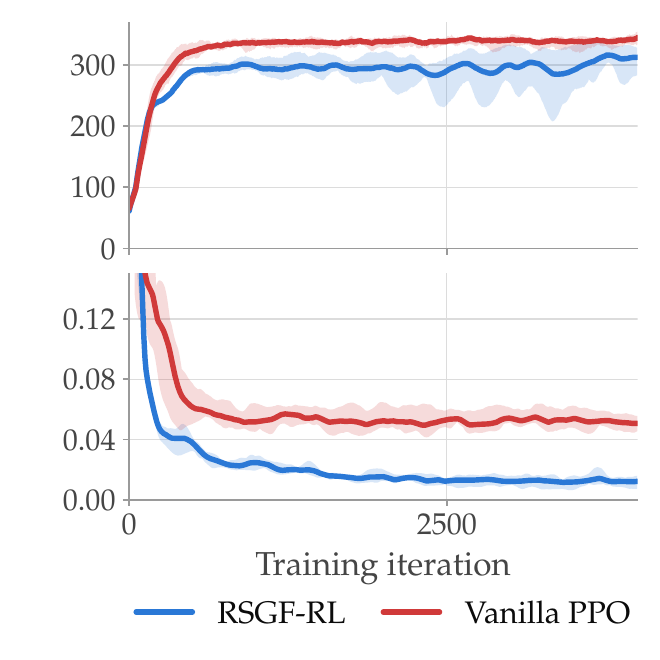}%
    \addrowpanel{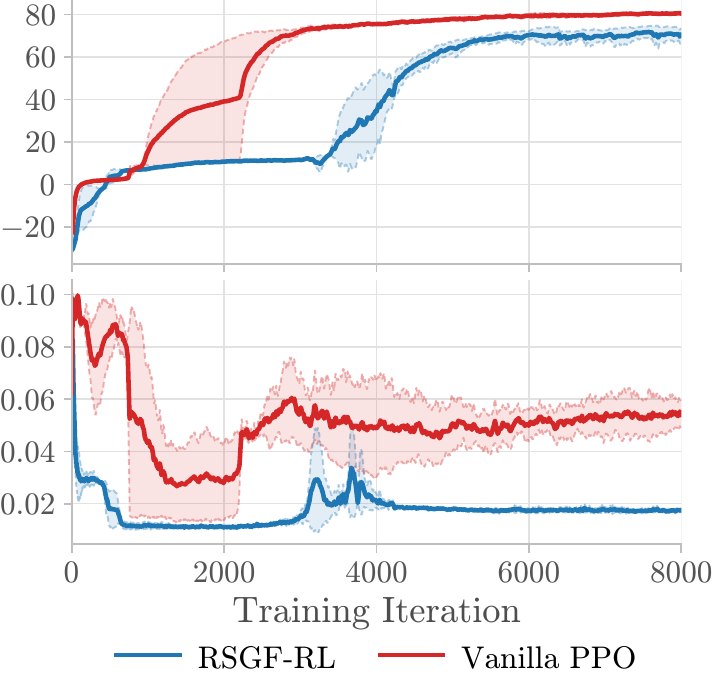}%
    \setlength{\rowH}{\dimexpr 0.97\linewidth*6553600/\rowsum\relax}%
    \subfigure[\scriptsize{Squatting}]{\label{fig:squatting}%
        \includegraphics[height=\rowH]{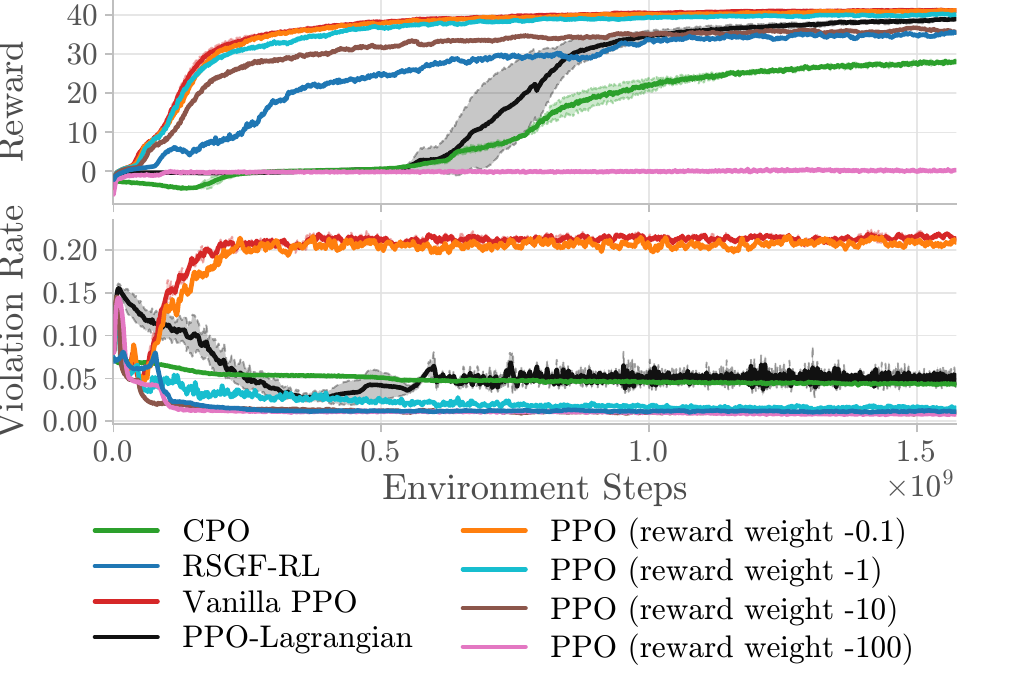}}\hfill
    \subfigure[\scriptsize{Crawling}]{\label{fig:crawling}%
        \includegraphics[height=\rowH]{figures/crawling6.pdf}}\hfill
    \subfigure[\scriptsize{Obstacle Avoid.}]{\label{fig:crazyflie}%
        \includegraphics[height=\rowH,trim=19.4bp 0 0 0,clip]{figures/crazy_safeppo_vs_ppo-3.pdf}}\hfill
    \subfigure[\scriptsize{Walking}]{\label{fig:walking}%
        \includegraphics[height=\rowH]{figures/walking5.pdf}}
    \vspace{-6pt}
    \caption{
    \small{
    Training curves for vanilla PPO and RSGF-RL. In each panel, the top plot shows the mean reward and the bottom plot the safety violation rate per episode (i.e., the fraction of episodes that contain at least one safety violation). 
    Shaded areas represent $\pm 1$ standard deviation over $5$ seeds.
    Panel~(a) also compares against CPO~\citep{JA-DH-AT-PA:17}, PPO-Lagrangian~\citep{ray2019benchmarking} and vanilla PPO with a safety-aware reward penalty for several penalty weights, and uses \textit{environment steps} on the horizontal axis since CPO uses a different number of steps per environment.}
    }
    \label{fig:training-comparison}
\end{figure}
We evaluate RSGF-RL on whole-body control tasks for a Unitree G1 humanoid robot, whose high dimensionality, contact-rich dynamics, and underactuation make safety-critical controller design challenging.
In all tasks, we train the RL policy in mjlab~\citep{KZ-QL-BY-LLL-KS-PA:26}. Simulated dynamics run at 200 Hz and the policy at 50 Hz, with 10 second episodes. The learned policies output joint position set-point commands that are tracked by a low-level PD controller. For sim-to-real transfer, we randomize base disturbances, contact friction, and center-of-mass and joint-encoder estimation errors.

First, we learn a fixed forward-gait walking policy, where the safe set requires a base height of at least $0.67 \text{ m}$, roll in $[-0.35 , 0.35] \text{ rad}$, pitch in $[-0.5 , 0.5 ]  \text{ rad}$, and the center of pressure (CoP) inside the support polygon of foot contact points.
%
Figure~\ref{fig:walking} shows that, when trained with the same $g_0$, RSGF-RL learns a policy with substantially fewer safety constraint violations than vanilla PPO, while asymptotically achieving comparable rewards.

Next, we test RSGF-RL on crawling and squatting motion imitation tasks (Figures~\ref{fig:crawling-hardware}-\ref{fig:squatting-hardware}), training a separate policy for each with motion imitation rewards~\citep{liao2026beyondmimic} and reference motions from a human motion dataset~\citep{BONES:25}.
We constrain the base height to be at least $0.4 \text{ m}$ for squatting, and the base and head heights to be at most $0.2 \text{ m}$ for crawling (with a curriculum from $0.3 \text{ m}$ to $0.2 \text{ m}$).
Figures~\ref{fig:squatting}-\ref{fig:crawling} show that, with the same $g_0$, RSGF-RL learns the squatting and crawling motions while respecting the height constraint, whereas vanilla PPO does not.
Figure~\ref{fig:squatting} also shows that for the squatting motion task, RSGF-RL improves upon CPO~\citep{JA-DH-AT-PA:17} both in terms of safety and performance, and upon PPO-Lagrangian~\citep{ray2019benchmarking} in safety, with very similar performance.
Additionally, vanilla PPO with an added safety-aware reward has safety and performance similar to RSGF-RL, as long as the penalty weight of the safety-aware reward is chosen appropriately. However, the correct penalty weight is very task-dependent, and it is unclear how to select it a priori.

\subsection{Drone collision avoidance}

\begin{figure}[t]
    \centering
    \renewcommand*{\subfigurelabel}[1]{{\scriptsize(\emph{#1})}}%
    \subfigure[\scriptsize{Crawling}]{\label{fig:crawling-hardware}%
        \begin{minipage}[b]{0.32\textwidth}\centering
            \includegraphics[width=\linewidth]{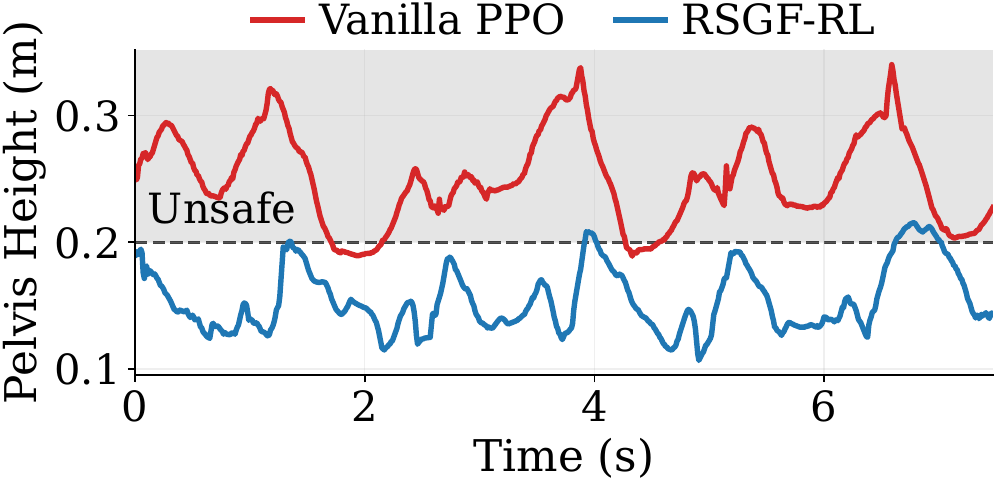}\\[0.02\linewidth]
            \parbox[b][0.5625\linewidth][t]{\linewidth}{\centering\includegraphics[height=0.5625\linewidth]{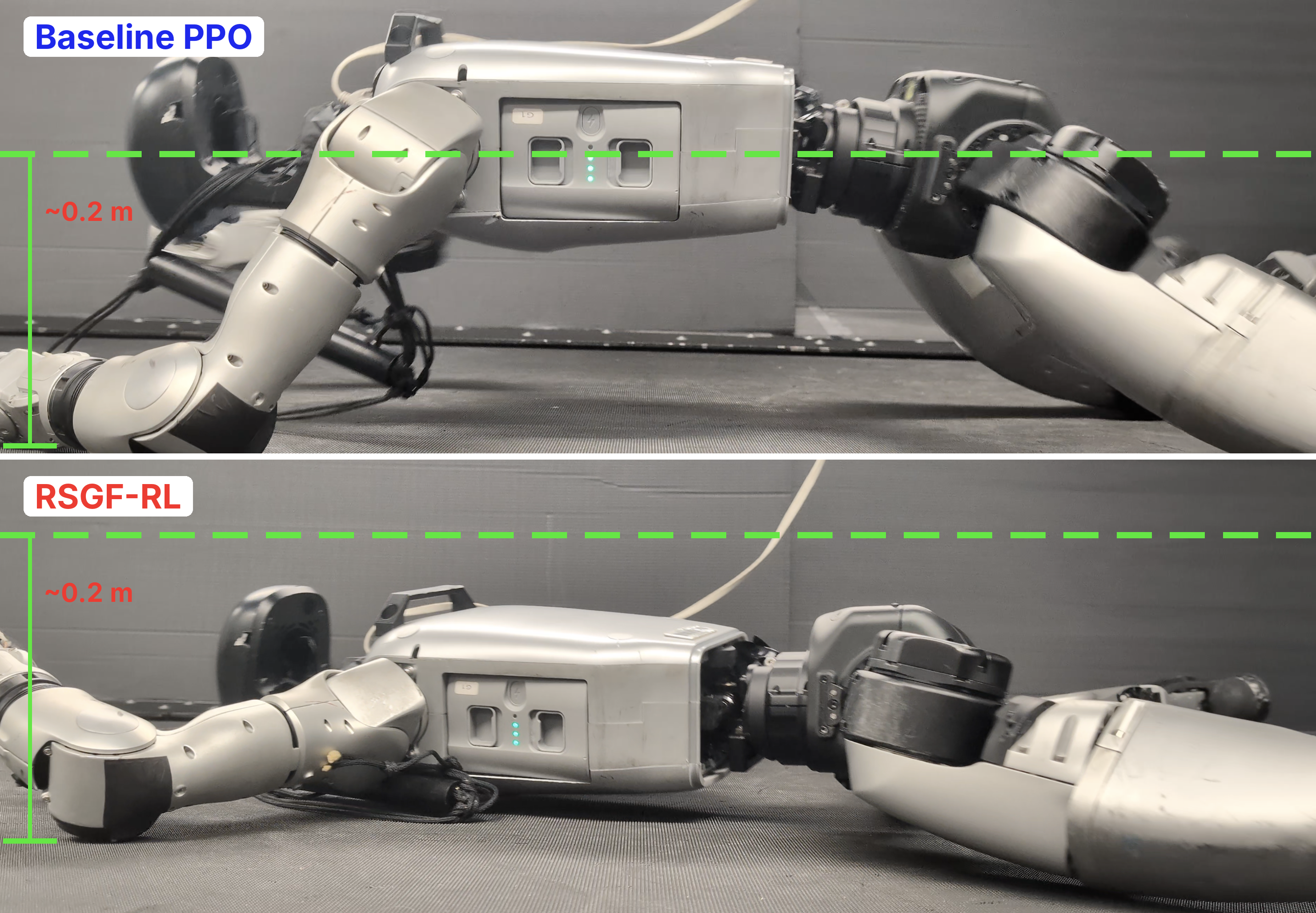}}%
        \end{minipage}}\hfill
    \subfigure[\scriptsize{Squatting}]{\label{fig:squatting-hardware}%
        \begin{minipage}[b]{0.32\textwidth}\centering
            \includegraphics[width=\linewidth]{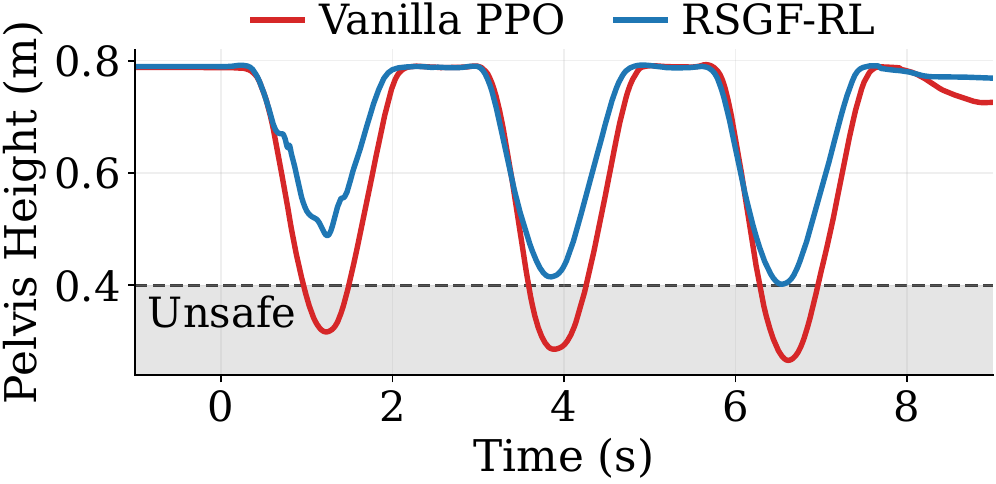}\\[0.02\linewidth]
            \parbox[b][0.5625\linewidth][t]{\linewidth}{\includegraphics[width=\linewidth]{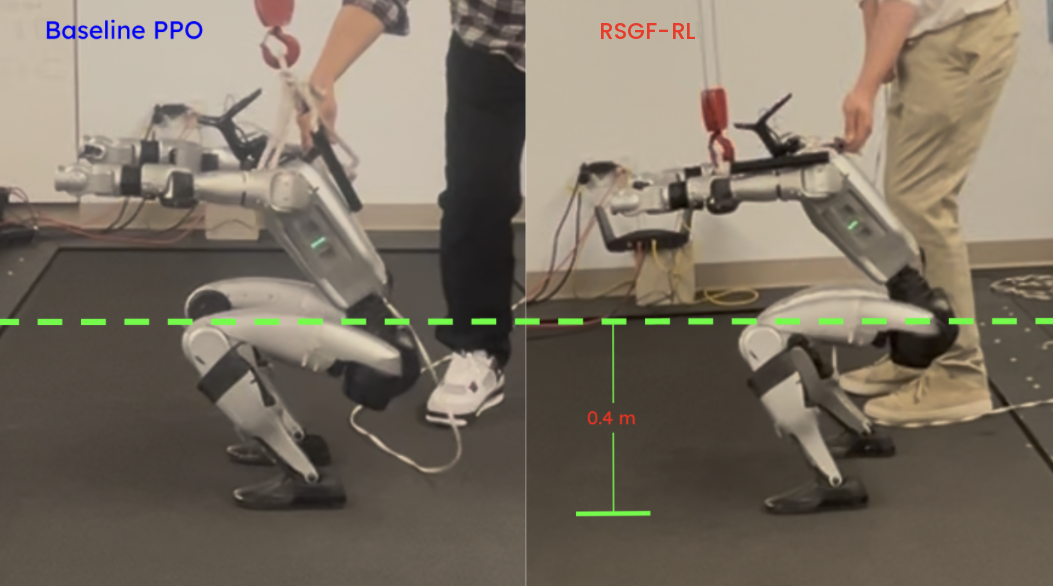}}%
        \end{minipage}}\hfill
    \subfigure[\scriptsize{Obstacle Avoidance}]{\label{fig:crazy_hardware}%
        \begin{minipage}[b]{0.32\textwidth}\centering
            \includegraphics[width=\linewidth]{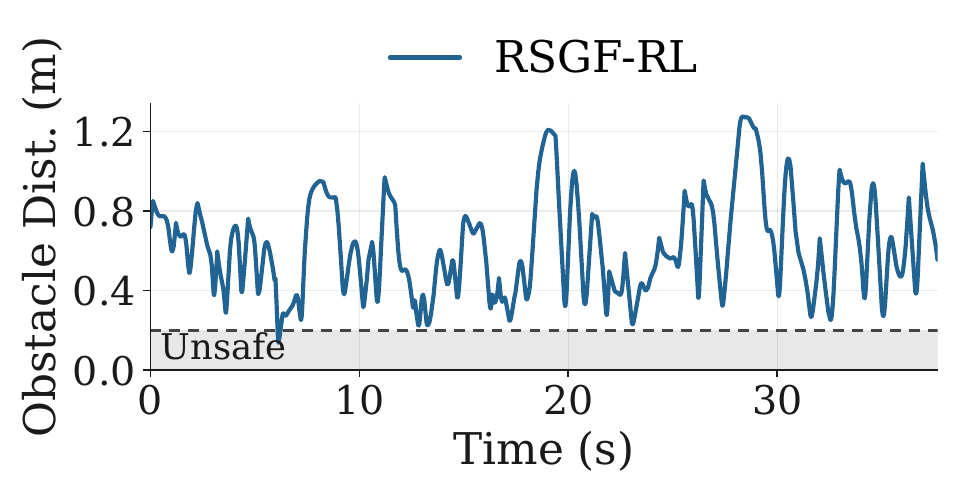}\\[0.02\linewidth]
            \parbox[b][0.5625\linewidth][t]{\linewidth}{\includegraphics[width=\linewidth]{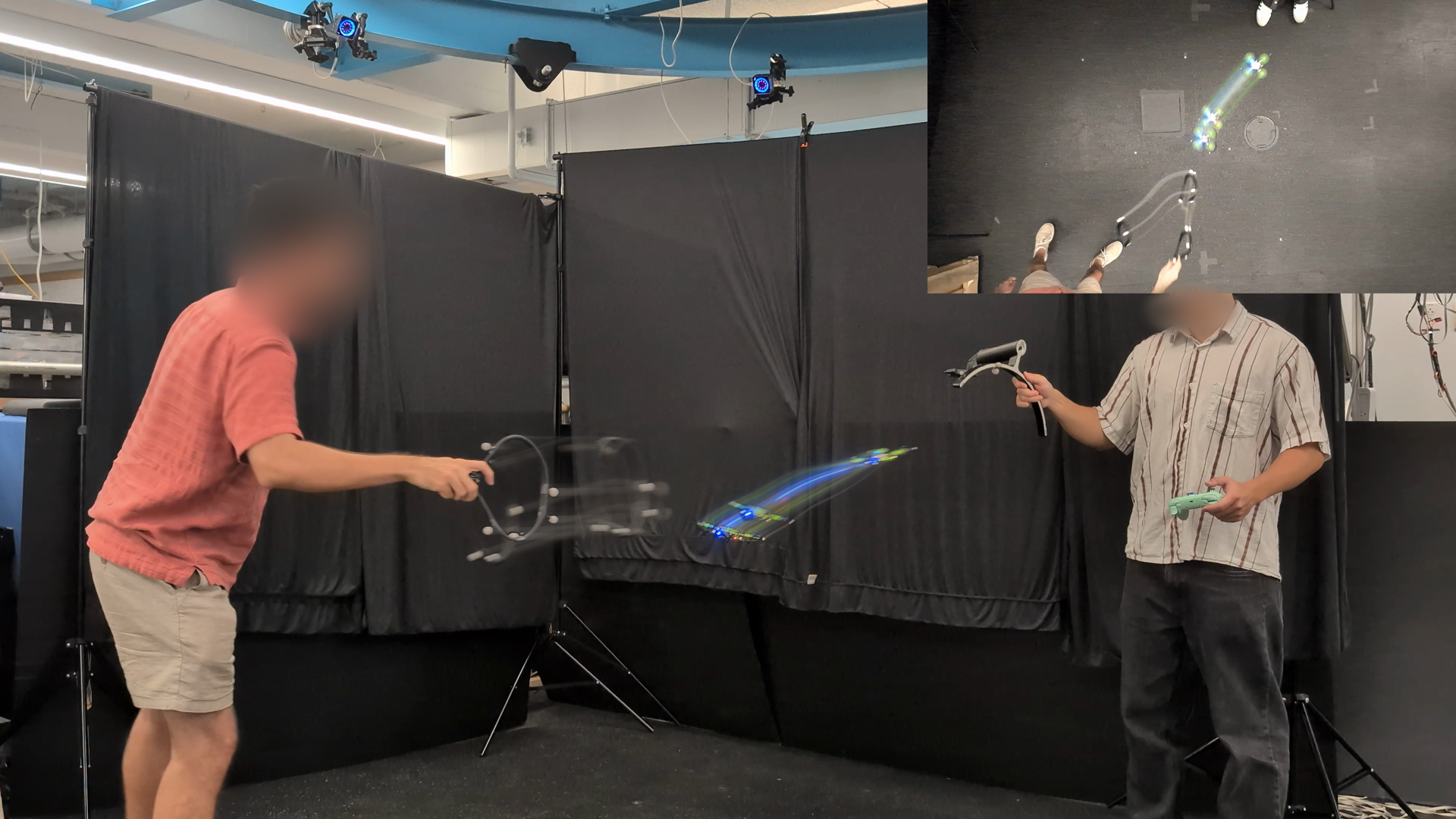}}%
        \end{minipage}}
    \vspace{-6pt}
    \caption{\small{(a)--(b) Humanoid crawling and squatting: vanilla PPO violates the safety constraints whereas RSGF-RL does not. (c) The RSGF-RL drone policy avoids two approaching obstacles.}}
    \vspace{-10pt}
    \label{fig:hardware}
\end{figure}

We learn a policy that steers a Crazyflie drone to a hover target that changes every $5$ seconds, in an arena with a randomized number of obstacles, some of which move randomly and others toward the drone. 
For sim-to-real transfer, we use domain randomization with random perturbations on the velocity, thrust coefficient, drag, command delays, and low-level controller feedback delay.
The learned policy outputs velocity commands that are tracked by the drone’s onboard low-level controller.
We train the RL policy in IsaacLab~\citep{mittal2025isaaclab}. The simulated dynamics run at $400$ Hz and the policy at $100$ Hz, with $10$ second episodes. 
In this case, we add safety penalties to vanilla PPO for entering, or being predicted to enter within 0.5 seconds, the safety margin of obstacles or the arena boundary, with a terminal penalty for collisions or leaving the arena. Figure~\ref{fig:crazyflie} again shows that RSGF-RL achieves a task return comparable to vanilla PPO while maintaining a low safety violation rate throughout training, whereas vanilla PPO sacrifices safety for higher return. Increasing these penalties made vanilla PPO training unstable. Although more careful tuning might improve its safety, RSGF-RL has no penalty parameter, and its own parameters ($\alpha, \beta, \{ h_t \}$) required little tuning in this experiment.
On hardware, the RSGF-RL policy successfully avoids obstacles while staying within the arena boundary (Fig.~\ref{fig:crazy_hardware}).

\subsection{Discussion on theory to practice gap}\label{sec:theory-practice-gap}

Some assumptions of our results, such as the bound~\eqref{eq:N-bound-safety} on the number of rollouts and the stepsize condition $0 < h_t \leq \bar{h}$, are not met in the experiments, and the probabilistic safety guarantees in Propositions~\ref{prop:safety-RSGF-RL} and~\ref{prop:cumulative-implies-probabilistic} are loose compared to the safety achieved in practice.

Moreover, our safety and convergence results depend on the estimator biases $\zeta_1, \tilde{\zeta}_0, \tilde{\zeta}_1$.
The estimates used in Section~\ref{sec:results}, detailed in Appendix~\ref{sec:training-parameters}, rely on standard
practical RL techniques, such as learned value critics, multiple Adam steps on the PPO loss, and omitting the
discount weighting 
of the policy gradient, each of which introduces bias. Although techniques such as
those in~\citep{LD-DS-MM-SK-BW-TA-SW:26},~\citep{HK-AK-AR:23} can be used to bound these biases, we do not have tight
characterizations for them. Finally, the curriculum in the crawling task makes the constraint iteration dependent, and
is not directly covered by our analysis. However, Proposition~\ref{prop:safety-RSGF-RL} still applies for
policy updates with the same constraint.

\section{Conclusions}

We have extended RSGF-RL to a general class of constraints, including probabilistic and cumulative ones, and to biased gradient and value function estimators, enabling its integration with state-of-the-art RL algorithms such as PPO.
We established safety and convergence guarantees for the resulting algorithm, and demonstrated it on whole-body motor skills for a G1 humanoid robot and collision-free waypoint navigation for a Crazyflie drone.
Future work will refine our assumptions and bias bounds and design estimators with good practical performance and provable guarantees, in order to narrow the gap between theory and practice discussed in Section~\ref{sec:theory-practice-gap}.

\acks{The authors used various AI assistants to write code, check arguments, and edit the paper. The authors verified all results in the paper and take full responsibility for the content.}

\bibliography{bib/alias,bib/Main-add,bib/Main,bib/JC,bib/PM,bib/New}

\appendix
\section{Connection between cumulative and probabilistic constraint}\label{sec:appendix-connection-cumulative-probabilistic}

Here we show that, with an appropriate choice of the cost function $R_1$, the cumulative
constraint $V_1(\btheta)\le 0$ induced by the choice of $g_1$ in Example~\ref{ex:cumulative-constraints}
implies (up to a term that vanishes as $\gamma\to1$) the probabilistic constraint in~\eqref{eq:probabilistic-constraint-problem}.

Consider
\begin{align}\label{eq:R1-choice-probabilistic}
    R_1(s_t,a_t,s_{t+1}) = -\mathbb{I}(s_t\in\bar{\Cc}) + 1 - \frac{\delta}{T+1},
\end{align}
which assigns a cost $1-\frac{\delta}{T+1}$ to unsafe states and $-\frac{\delta}{T+1}$ to safe ones.
In practice, we use a smoothened version of~\eqref{eq:R1-choice-probabilistic} given by
\begin{align}\label{eq:R1-practical-choice-probabilistic-safety}
    R_1(s_t,a_t,s_{t+1}) =
    \begin{cases}
        \frac{\delta}{T+1}\big( e^{-\tilde{h}(s_t)} - 1\big), \quad &\text{if} \ s_t\in\bar{\Cc}, \\
        1 - \frac{\delta}{T+1}, \quad &\text{else},
    \end{cases}
\end{align}
where $\tilde{h}:\mathcal{S}\to\real$ is a function whose $0$-superlevel set includes $\bar{\Cc}$.
Since $\tilde{h}(s_t)\ge 0$ for $s_t\in\bar{\Cc}$, \eqref{eq:R1-practical-choice-probabilistic-safety}
upper bounds~\eqref{eq:R1-choice-probabilistic} pointwise, while additionally encouraging the
state to stay away from the boundary of $\bar{\Cc}$.
For $\gamma\in(0,1]$, define
\begin{align}\label{eq:kappa-gamma}
    \kappa(\gamma) = \frac{1}{T+1}\sum_{k=0}^{T}\gamma^{-k}
    = \frac{1-\gamma^{T+1}}{(T+1)(1-\gamma)\gamma^{T}} \quad (\gamma<1),
\end{align}
$\kappa(1) = 1$, and $\kappa(\gamma)\ge 1$ for all $\gamma\in(0,1]$.

\begin{proposition}\label{prop:cumulative-implies-probabilistic}
Let $\gamma\in(0,1]$, let $g_1$ be as in Example~\ref{ex:cumulative-constraints}, and let $R_1$ be
either~\eqref{eq:R1-choice-probabilistic} or~\eqref{eq:R1-practical-choice-probabilistic-safety}.
If $V_1(\btheta)\le 0$, then
\begin{align}\label{eq:prob-safety-gamma}
    \mathbb{P}\Big(\bigcap_{t=0}^T\{s_t\in\bar{\Cc}\}\Big) \ge 1-\delta-\varepsilon(\gamma),
    \qquad
    \varepsilon(\gamma) = \delta\big(\kappa(\gamma)-1\big) \le \frac{\delta\,T(1-\gamma)}{2\gamma^{T}}.
\end{align}
In particular, for $\gamma=1$ the probabilistic constraint in~\eqref{eq:probabilistic-constraint-problem}
holds exactly, and for fixed $T$ its violation vanishes linearly in $1-\gamma$ as $\gamma\to1$.
\end{proposition}
\begin{proof}
Let $U_t=\mathbb{I}(s_t\notin\bar{\Cc})$, so that~\eqref{eq:R1-choice-probabilistic} reads
$R_1 = U_t-\frac{\delta}{T+1}$. Since~\eqref{eq:R1-practical-choice-probabilistic-safety}
upper bounds~\eqref{eq:R1-choice-probabilistic}, in both cases
\begin{align}\label{eq:V1-Ut-bound}
    0 \ge V_1(\btheta) \ge \mathbb{E}\Big[\sum_{t=0}^T\gamma^t U_t\Big] - \frac{\delta}{T+1}\sum_{t=0}^T\gamma^t .
\end{align}
Using the union bound on probabilities~\cite[Theorem 1.1.1]{RD:10-5th-edition} 
and $\gamma^{t}\ge\gamma^{T}$ for $t\le T$,
\begin{align*}
    \mathbb{P}\Big(\bigcup_{t=0}^T\{s_t\notin\bar{\Cc}\}\Big)
    \le \mathbb{E}\Big[\sum_{t=0}^T U_t\Big]
    \le \gamma^{-T}\,\mathbb{E}\Big[\sum_{t=0}^T\gamma^t U_t\Big]
    \le \frac{\delta}{T+1}\sum_{t=0}^T\gamma^{t-T}
    = \delta\,\kappa(\gamma),
\end{align*}
where in the last inequality we used~\eqref{eq:V1-Ut-bound}, and gives the first part of~\eqref{eq:prob-safety-gamma}. For the bound on $\varepsilon(\gamma)$,
note that $1-\gamma^k = (1-\gamma)\sum_{i=0}^{k-1}\gamma^i \le k(1-\gamma)$ and $\gamma^k\ge\gamma^T$, so
$\gamma^{-k}-1 = \frac{1-\gamma^k}{\gamma^k} \le \frac{k(1-\gamma)}{\gamma^T}$. Hence
\begin{align*}
    \kappa(\gamma)-1 = \frac{1}{T+1}\sum_{k=0}^T\big(\gamma^{-k}-1\big)
    \le \frac{1-\gamma}{(T+1)\gamma^T}\sum_{k=0}^T k = \frac{T(1-\gamma)}{2\gamma^T}. \qquad
\end{align*}
\end{proof}

\paragraph{Remarks.}
The case $\gamma=1$ with~\eqref{eq:R1-choice-probabilistic} was established in~\citep[Theorem 1]{SP-MCF-LFOC-AR:23}.
However, in our experiments we found that taking $\gamma<1$, together with the choice of $R_1$
in~\eqref{eq:R1-practical-choice-probabilistic-safety}, leads to better practical performance.
Proposition~\ref{prop:cumulative-implies-probabilistic} shows that the price paid for this choice is
a violation of the probabilistic constraint of at most $\varepsilon(\gamma)=\mathcal{O}\big(\delta T(1-\gamma)\big)$.
If exact satisfaction is required for $\gamma<1$, it suffices to replace the constant
$\frac{\delta}{T+1}$ in~\eqref{eq:R1-choice-probabilistic} and~\eqref{eq:R1-practical-choice-probabilistic-safety}
by $\frac{\delta}{T+1}\kappa(\gamma)^{-1} = \frac{\delta\gamma^T(1-\gamma)}{1-\gamma^{T+1}}$,
at the cost of a more conservative constraint.
Finally, since Proposition~\ref{prop:cumulative-implies-probabilistic} only provides a sufficient
condition for the satisfaction of the probabilistic constraint~\eqref{eq:probabilistic-constraint-problem},
(assuming $\gamma = 1$ or correct constraint tightening), the two formulations are not equivalent.

\section{Properties of the RSGF}\label{sec:details-proof-Proposition-RSGF-properties}

We next summarize the properties of~\eqref{eq:discrete-time-rsgf}, adapted from~\citep[Lemma 1]{PM-AM-JC:25-l4dc} and~\citep[Lemma IV.2, Prop. IV.3]{PM-AM-JC:25-arxiv}.

\begin{proposition}\longthmtitle{RSGF Properties}\label{prop:rsgf-properties}
    Suppose that~\eqref{eq:V0-V1-optimization-problem} satisfies the Mangasarian--Fromovitz constraint qualification (MFCQ)\footnote{For the single-constraint problem considered here, MFCQ requires that, for every $\btheta\in\Cc$ satisfying $V_1(\btheta)=0$, there exists $\bq\in\real^d$ such that $\nabla V_1(\btheta)^\top\bq<0$. Equivalently, $\nabla V_1(\btheta)\neq 0$ whenever $V_1(\btheta)=0$.} at every $\btheta\in\Cc$.
    Then~\eqref{eq:R-definition} is well-defined on an open neighborhood of $\Cc$.
    Moreover, $\btheta\in\Cc$ is a KKT point of~\eqref{eq:V0-V1-optimization-problem} if and only if $\Rc(\btheta)=0$.
    Furthermore, suppose that $\nabla V_1$ is $L_1$-Lipschitz and that
    $h_t < \min\{ 1/\alpha, \beta/L_1  \}$ for all $t\in\mathbb{N}$.
    Then $\Cc$ is forward invariant under~\eqref{eq:discrete-time-rsgf}.
    %
    If, in addition, $\nabla V_0$ is $L_0$-Lipschitz and
    $h_t<1/L_0$, then every limit point of a sequence
    $\{\btheta_t\}_{t\in\mathbb{N}}$ initialized at $\btheta_0\in\Cc$
    is a KKT point of~\eqref{eq:V0-V1-optimization-problem}.
\end{proposition}

Proposition~\ref{prop:rsgf-properties} is a direct consequence of~\citep[Lemma 1]{PM-AM-JC:25-l4dc} and~\citep[Lemma IV.2]{PM-AM-JC:25-arxiv}. However, we note that~\citep[Lemma IV.2]{PM-AM-JC:25-arxiv} requires that~\eqref{eq:R-definition} satisfies the Constant Rank Condition (CRC) at $(\btheta,\Rc(\btheta))$ for every $\btheta\in\Cc$, but that hypothesis is not stated in the statement of Proposition~\ref{prop:rsgf-properties}.

The following result explains the reason behind this mismatch: given that~\eqref{eq:V0-V1-optimization-problem} only has a single constraint, the satisfaction of MFCQ for~\eqref{eq:V0-V1-optimization-problem} at every $\btheta\in\Cc$ implies the satisfaction of CRC for~\eqref{eq:R-definition} at $(\btheta,\Rc(\btheta))$ for every $\btheta\in\Cc$.

\begin{lemma}\longthmtitle{MFCQ implies CRC for single constraint problems}\label{lem:mfcq-implies-crc-for-single-constraint-problems}
    Suppose that~\eqref{eq:V0-V1-optimization-problem} satisfies MFCQ for each $\btheta\in\Cc$.
    Then,~\eqref{eq:R-definition} satisfies CRC at $(\btheta,\Rc(\btheta))$ for every $\btheta\in\Cc$.
\end{lemma}
\begin{proof}
    Suppose that CRC does not hold for~\eqref{eq:R-definition} at some $\btheta\in\Cc$.
    Then we necessarily have that
    \begin{align}\label{eq:active-V1-constraint}
        \alpha V_1(\btheta) + \nabla V_1(\btheta)^\top \Rc(\btheta) + \frac{\beta}{2}\norm{\Rc(\btheta)}^2 = 0,
    \end{align}
    and 
    \begin{align}\label{eq:zero-gradient}
        \frac{\partial}{\partial\bxi}\Big(
        \alpha V_1(\btheta) + \nabla V_1(\btheta)^\top \xi + \frac{\beta}{2}\norm{\xi}^2
        \Big)\Bigg|_{\bxi = \Rc(\btheta)} = \nabla V_1(\btheta) + \beta \Rc(\btheta) = 0.
    \end{align}
    Plugging~\eqref{eq:zero-gradient} into~\eqref{eq:active-V1-constraint} we get 
    \begin{align}\label{eq:CRC-conclusion}
        \alpha V_1(\btheta) - \frac{\beta}{2}\norm{\Rc(\btheta)}^2 = 0.
    \end{align}
    Since $\btheta\in\Cc$, we have $V_1(\btheta) \leq 0$. Hence,~\eqref{eq:CRC-conclusion} implies that $V_1(\btheta) = 0$ and $\Rc(\btheta) = 0$, which from~\eqref{eq:zero-gradient} implies that $\nabla V_1(\btheta) = 0$, violating MFCQ for~\eqref{eq:V0-V1-optimization-problem} at $\btheta\in\Cc$.
    Hence, CRC necessarily holds for~\eqref{eq:R-definition} at $(\btheta,\Rc(\btheta))$ for every $\btheta\in\Cc$.
\end{proof}

\section{Proof of Proposition~\ref{prop:policy-gradient-general}}\label{sec:appendix-proof-policy-gradient-general}

When $g_j$ is a discounted cumulative return, as in Example~\ref{ex:cumulative-constraints}, and $b_{j,t}\equiv0$, Proposition~\ref{prop:policy-gradient-general} recovers the standard Policy Gradient Theorem~\citep[Section 13.2]{RSS-AGB:18}. If the baseline is time-independent and depends only on $s_t$, it recovers the corresponding baseline formulation~\citep[Section 13.4]{RSS-AGB:18}. Likewise, for the probabilistic constraint in Example~\ref{ex:probabilistic-constraints} with $b_{j,t}\equiv0$, it reduces to~\citep[Theorem 2]{WC-DS-SP:24}. As usual, the baselines $\{b_{j,t}\}$ leave the expected gradient unchanged (as shown next in the proof) while potentially reducing its variance.

\begin{proof}
    The proof is standard and follows arguments analogous to~\citep{PWG:90} and the Policy Gradient Theorem~\citep[Chapter 13]{RSS-AGB:18}. We include it here for completeness.
    Let us first show that
    \begin{align}\label{eq:first-policy-gradient-expression}
        \nabla V_j(\btheta) = \mathbb{E}\Big[ 
        \sum_{t=0}^T g_j(\bs,\ba) \nabla_{\btheta} \log\pi_{\btheta}(a_t|s_t)
        \Big].
    \end{align}
    Let $\sigma:\Sc^{T+2}\times\Ac^{T+1}\times\real^d\to\real$ be defined as
    \begin{align*}
        \sigma(\bs,\ba,\btheta) = \eta(s_0)\prod_{t=0}^T \pi_{\btheta}(a_t|s_t)P(s_{t+1}|a_t,s_t).
    \end{align*}
    Since the hypothesis for differentiation under the integral sign hold~\citep[Theorem 10.39]{TMA:74} (because $\btheta\to\log\pi_{\btheta}(a|s)$ is continuously differentiable and for each $\btheta\in\real^d$,~\eqref{eq:uniform-boundedness-logpitheta} holds), we have
    \begin{align*}
        \nabla V_j(\btheta) = \int_{ \Gamma } g_j(\bs,\ba) \sigma(\bs,\ba,\btheta) \nabla_{\btheta}\log\sigma(\bs,\ba,\btheta) d\bs d\ba,
    \end{align*}
    where $\Gamma = \Sc^{T+2}\times\Ac^{T+1}$.
    Now, note that since 
    \begin{align*}
        \log \sigma(\bs,\ba,\btheta) = \log \eta(s_0) + \sum_{t=0}^T \! \Big( 
        \log \! P(s_{t+1}|a_t,s_t) 
        +
        \log\pi_{\btheta}(a_t|s_t)
        \Big),
    \end{align*}
    we have 
    \begin{align*}
        \nabla V_j(\btheta) &= \int_{ \Gamma }  g_j(\bs,\ba) \sigma(\bs,\ba,\btheta) \sum_{t=0}^T \nabla_{\btheta}\log\pi_{\btheta}(a_t|s_t) d\bs d\ba,
    \end{align*}
    proving~\eqref{eq:first-policy-gradient-expression}.
    Next, let us show that for each $t\in0:T$, we have 
    \begin{align*}
        \mathbb{E}[b_{j,t}(H_t,\btheta)\nabla_{\btheta}\log\pi_{\btheta}(a_t|s_t)] = 0,
    \end{align*}
    which will complete the proof.
    Using the tower property of conditional expectation~\cite[Section 9.7]{DW:91}, 
    \begin{align*}
        &\mathbb{E}[b(H_t,\btheta)\nabla_{\btheta}\log\pi_{\btheta}(a_t|s_t) ] = \mathbb{E}_{H_t}[ \mathbb{E}_{a_t}[b(H_t,\btheta)\nabla_{\btheta}\log\pi_{\btheta}(a_t|s_t) | H_t ] ],
    \end{align*}
    where on the right hand side expression, the outer expectation is with respect to the history $H_t$ and the inner expectation is with respect to action $a_t$ only.
    Now, note that for any fixed $H_t$, since $H_t$ does not depend on $a_t$ and $a_t \sim \pi_{\btheta}(\cdot|s_t)$,
    we have
    \begin{align*}
        \mathbb{E}_{a_t}[ b(H_t,\btheta) \nabla_{\btheta} \log\pi_{\btheta}(a_t|s_t) | H_t ] &= b(H_t,\btheta) \int_{\mathcal{A}} \nabla_{\btheta} \log \pi_{\btheta}(a_t|s_t) \pi_{\btheta}(a_t|s_t) da_t \\
        &=b(H_t,\btheta) \int_{\mathcal{A}} \nabla_{\btheta} \pi_{\btheta}(a_t|s_t) da_t = 0,
    \end{align*}
    where in the last equality we have used the fact that since $1 = \int_{\mathcal{A}} \pi_{\btheta}(a_t|s_t) da_t$, by differentiating under the integral sign we have $0 = \int_{\mathcal{A}} \nabla_{\btheta} \pi_{\btheta}(a_t|s_t) da_t$.
\end{proof}

\section{Statistical properties of estimators}\label{sec:statistical-properties-estimators}

The following result establishes different statistical properties of the estimators~\eqref{eq:general-estimates}, which are needed for the proof of Proposition~\ref{prop:safety-RSGF-RL}.

\begin{proposition}\longthmtitle{Statistical properties of estimators}\label{prop:statistical-properties}
    Let $j\in\{0,1\}$, $\btheta\in\real^d$, and $\widehat{V}_j(\btheta)$, $\widehat{\nabla V}_j(\btheta)$ be defined as in~\eqref{eq:general-estimates}, and suppose that conditions~\eqref{eq:boundedness-estimates},~\eqref{eq:bounded-bias} hold.
    Then,
    \begin{align*}
        &|\mathbb{E}[\widehat{V}_j(\btheta)] - V_j(\btheta)| \leq \zeta_j, \ 
        \norm{ \mathbb{E}[\widehat{\nabla V}_j(\btheta)] - \nabla V_j(\btheta) } \leq \sqrt{d} \tilde{\zeta}_j,\\
        &\mathbb{P}\Big( 
        |\widehat{V}_j(\btheta) - V_j(\btheta)| \leq \zeta_j + \epsilon
        \Big) \geq 1 - 2\exp\Big\{ -\frac{N\epsilon^2}{2 B_j^2} \Big\}, \\
        &\mathbb{P}\Big( 
        \norm{ \widehat{\nabla V}_j(\btheta) - \nabla V_j(\btheta) }
        \leq \sqrt{d}\tilde{\zeta}_j + \epsilon 
        \Big) 
        \geq 1 - 2d\exp\Big\{ 
        -\frac{N \epsilon^2}{2d\tilde{B}_j^2}
        \Big\},
    \end{align*}
    for all $\epsilon>0$.
\end{proposition}
\begin{proof}
The proof of Proposition~\ref{prop:statistical-properties} follows an argument analogous to that of~\citep[Lemma 2]{PM-AM-JC:25-l4dc}, but we include it here for completeness.
Fix $j\in\{0,1\}$ and $\btheta\in\mathbb{R}^d$.
We first establish the bounds on the bias. By definition of $\widehat V_j(\btheta)$
and condition~(9),
\begin{align*}
\left|\mathbb{E}[\widehat V_j(\btheta)]-V_j(\btheta)\right|
=
\left|
\frac{1}{N}\sum_{n=1}^N
\left(\mathbb{E}[G_{j,n}^{\btheta}]-V_j(\btheta)\right)
\right|
\le
\frac{1}{N}\sum_{n=1}^N
\left|\mathbb{E}[G_{j,n}^{\btheta}]-V_j(\btheta)\right|
\le \zeta_j.
\end{align*}
Similarly, for every $i\in 1:d$,
\begin{align*}
\left|
\left(\mathbb{E}[\widehat{\nabla V_j}(\btheta)]
-\nabla V_j(\btheta)\right)_i
\right|
=
\left|
\frac{1}{N}\sum_{n=1}^N
\left(
\mathbb{E}[(\tilde G_{j,n}^{\btheta})_i]
-(\nabla V_j(\btheta))_i
\right)
\right| \le \tilde\zeta_j.
\end{align*}
Consequently,
\[
\left\|
\mathbb{E}[\widehat{\nabla V_j}(\btheta)]
-\nabla V_j(\btheta)
\right\|
\le
\sqrt{\sum_{i=1}^d\tilde\zeta_j^2}
=
\sqrt d\,\tilde\zeta_j.
\]

We next establish the concentration bounds. Since the $N$ episodes are
independent, the random variables $\{G_{j,n}^{\btheta}\}_{n=1}^N$ are
independent. Moreover, by~(10),
$G_{j,n}^{\btheta}\in[-B_j,B_j]$ almost surely. Hence, Hoeffding's
inequality~\citep{WH:63-hoeffding} yields, for every $\epsilon>0$,
\[
\mathbb{P}\left(
\left|
\widehat V_j(\btheta)-\mathbb{E}[\widehat V_j(\btheta)]
\right|>\epsilon
\right)
\le
2\exp\left\{
-\frac{N\epsilon^2}{2B_j^2}
\right\}.
\]
Using
\begin{align*}
\left|\widehat V_j(\btheta)-V_j(\btheta)\right|
\le
\left|\widehat V_j(\btheta)
-\mathbb{E}[\widehat V_j(\btheta)]\right|
+
\left|\mathbb{E}[\widehat V_j(\btheta)]
-V_j(\btheta)\right|
\le
\left|\widehat V_j(\btheta)
-\mathbb{E}[\widehat V_j(\btheta)]\right|
+\zeta_j,
\end{align*}
we obtain
\[
\mathbb{P}\left(
\left|\widehat V_j(\btheta)-V_j(\btheta)\right|
\le \zeta_j+\epsilon
\right)
\ge
1-
2\exp\left\{
-\frac{N\epsilon^2}{2B_j^2}
\right\}.
\]

We finally consider the gradient estimator. For every $i\in 1:d$,
condition~(10) implies
$(\widetilde G_{j,n}^{\btheta})_i
\in[-\tilde{B}_j,\tilde{B}_j]$ almost surely.
Thus, Hoeffding's inequality~\citep{WH:63-hoeffding} gives
\[
\mathbb{P}\left(
\left|
(\widehat{\nabla V_j}(\btheta))_i
-
(\mathbb{E}[\widehat{\nabla V_j}(\btheta)])_i
\right|
>
\frac{\epsilon}{\sqrt d}
\right)
\le
2\exp\left\{
-\frac{N\epsilon^2}
{2d \tilde{B}_j^2}
\right\}.
\]
Therefore, by Fr\'echet's inequality~\citep{MF:35},
\begin{align}\label{eq:gradient-hoeffding-frechet}
\mathbb{P}\left(
\bigcap_{i=1}^d
\left\{
\left|
(\widehat{\nabla V_j}(\btheta))_i
-
(\mathbb{E}[\widehat{\nabla V_j}(\btheta)])_i
\right|
\le
\frac{\epsilon}{\sqrt d}
\right\}
\right)
\ge
1-
2d\exp\left\{
-\frac{N\epsilon^2}
{2d \tilde{B}_j^2}
\right\}.
\end{align}
Now, if the event inside the probability in~\eqref{eq:gradient-hoeffding-frechet} holds, we have that
\[
\left\|
\widehat{\nabla V_j}(\btheta)
-
\mathbb{E}[\widehat{\nabla V_j}(\btheta)]
\right\|
\le
\sqrt{
\sum_{i=1}^d
\frac{\epsilon^2}{d}}
=\epsilon.
\]
Combining this inequality with the bias bound established above gives
\begin{align*}
\left\|
\widehat{\nabla V_j}(\btheta)-\nabla V_j(\btheta)
\right\|
\le
\left\|
\widehat{\nabla V_j}(\btheta)
-
\mathbb{E}[\widehat{\nabla V_j}(\btheta)]
\right\|
+
\left\|
\mathbb{E}[\widehat{\nabla V_j}(\btheta)]
-\nabla V_j(\btheta)
\right\| \leq
\epsilon+\sqrt d\,\tilde{\zeta}_j.
\end{align*}
Hence,
\[
\mathbb{P}\left(
\left\|
\widehat{\nabla V_j}(\btheta)-\nabla V_j(\btheta)
\right\|
\le
\sqrt d\,\tilde{\zeta}_j+\epsilon
\right)
\ge
1-
2d\exp\left\{
-\frac{N\epsilon^2}
{2d \tilde{B}_j^2}
\right\},
\]
which completes the proof.
\end{proof}

\section{Proof of Proposition~\ref{prop:safety-RSGF-RL}}\label{sec:proof-proposition-safety-RSGF-RL}

\begin{proof}
Throughout, fix $t\in 0:K-1$ and let
$\Delta_t := \btheta_{t+1}-\btheta_t = h_t\hat{\mathcal{R}}(\btheta_t)$.
 
\smallskip\noindent\emph{Step 1 (one-step safety evolution).}
Since $\nabla V_1$ is $L_1$-Lipschitz, by~\cite[Lemma 1.2.3]{YN:18} we have
\begin{equation*}
V_1(\btheta_{t+1}) \le V_1(\btheta_t)
+ \nabla V_1(\btheta_t)^\top \Delta_t + \frac{L_1}{2}\|\Delta_t\|^2 .
\end{equation*}
Adding and subtracting $\widehat{\nabla V}_1(\btheta_t)^\top\Delta_t$ and using the
Cauchy--Schwarz inequality,
\begin{equation*}
V_1(\btheta_{t+1}) \le V_1(\btheta_t)
+ \widehat{\nabla V}_1(\btheta_t)^\top\Delta_t
+ \big\|\nabla V_1(\btheta_t)-\widehat{\nabla V}_1(\btheta_t)\big\|\,\|\Delta_t\|
+ \frac{L_1}{2}\|\Delta_t\|^2 .
\end{equation*}
Since~\eqref{eq:R-hat-definition} is feasible at $\btheta_t$, $\hat{\mathcal{R}}(\btheta_t)$ satisfies its constraint;
multiplying it by $h_t>0$ and using $\Delta_t=h_t\hat{\mathcal{R}}(\btheta_t)$ gives
$\widehat{\nabla V}_1(\btheta_t)^\top\Delta_t
\le -\alpha h_t\widehat{V}_1(\btheta_t)-\frac{\beta}{2h_t}\|\Delta_t\|^2$.
Using also the identity
$V_1(\btheta_t)-\alpha h_t\widehat{V}_1(\btheta_t)=(1-\alpha h_t)V_1(\btheta_t)-\alpha h_t\big(\widehat{V}_1(\btheta_t)-V_1(\btheta_t)\big)$,
we obtain
\begin{align}
V_1(\btheta_{t+1}) \le{}& (1-\alpha h_t)V_1(\btheta_t)
+ \alpha h_t\big|\widehat{V}_1(\btheta_t)-V_1(\btheta_t)\big| \nonumber\\
&+ \big\|\nabla V_1(\btheta_t)-\widehat{\nabla V}_1(\btheta_t)\big\|\,\|\Delta_t\|
- \frac{1}{2}\Big(\frac{\beta}{h_t}-L_1\Big)\|\Delta_t\|^2 . \label{eq:pf-step1}
\end{align}
Inequality~\eqref{eq:pf-step1} holds deterministically, i.e., for every
realization of the $N$ rollouts collected at iteration $t$.
 
\smallskip\noindent\emph{Step 2 (eliminating the step length).}
Since $h_t<\beta/L_1$, the coefficient $c_t:=\frac12(\beta/h_t-L_1)$ is positive. For every $e\ge0$, we have $\max_{x\ge0}\{ex-c_tx^2\}=e^2/(4c_t)$. Hence the last two terms of~\eqref{eq:pf-step1} are at most
$h_t\|\nabla V_1(\btheta_t)-\widehat{\nabla V}_1(\btheta_t)\|^2/(2(\beta-L_1h_t))$, and
\begin{equation}\label{eq:pf-step2}
V_1(\btheta_{t+1}) \le (1-\alpha h_t)V_1(\btheta_t)
+ \alpha h_t\big|\widehat{V}_1(\btheta_t)-V_1(\btheta_t)\big|
+ \frac{h_t\big\|\nabla V_1(\btheta_t)-\widehat{\nabla V}_1(\btheta_t)\big\|^2}{2(\beta-L_1h_{\max})},
\end{equation}
where we also used $h_t\le h_{\max}$.
 
\smallskip\noindent\emph{Step 3 (bounding estimates deviation from true values).}
Define the events
\begin{align*}
\mathcal{E}_{V,t} &:= \Big\{\big|\widehat{V}_1(\btheta_t)-V_1(\btheta_t)\big|\le \zeta_1+\epsilon\Big\},\\
\mathcal{E}_{\nabla,t} &:= \Big\{\big\|\widehat{\nabla V}_1(\btheta_t)-\nabla V_1(\btheta_t)\big\|\le \sqrt{d}\,\tilde\zeta_1+\epsilon\Big\},
\end{align*}
and $\mathcal{E}_t:=\mathcal{E}_{V,t}\cap\mathcal{E}_{\nabla,t}$.
Note that $\epsilon>0$ is fixed before sampling. Moreover,
$\btheta_t$ is $\mathcal{F}_t$-measurable and, conditionally on
$\mathcal{F}_t$, the $N$ episodes collected at iteration $t$ are independent and
generated under $\pi_{\btheta_t}$. Hence Proposition~\ref{prop:statistical-properties} applies
conditionally on $\mathcal{F}_t$ with $\btheta=\btheta_t$, and yields
\begin{align*}
\mathbb{P}\big(\mathcal{E}_{V,t} \,\big|\, \mathcal{F}_t\big)
&\ge 1-2\exp\Big\{-\frac{N\epsilon^2}{2B_1^2}\Big\} \ge 1-\frac{\bar\delta}{2},\\
\mathbb{P}\big(\mathcal{E}_{\nabla,t} \,\big|\, \mathcal{F}_t\big)
&\ge 1-2d\exp\Big\{-\frac{N\epsilon^2}{2d\tilde B_1^2}\Big\} \ge 1-\frac{\bar\delta}{2},
\end{align*}
where the second inequality in each line follows from~\eqref{eq:N-bound-safety}.
By Fr\'echet's inequality~\citep{MF:35},
$\mathbb{P}(\mathcal{E}_t\mid\mathcal{F}_t)\ge 1-\bar\delta$.
By the union bound (cf.~\citealp[Theorem 1.1.1]{RD:10-5th-edition}) and the tower property (cf.~\citealp[Section 9.7]{DW:91}),
\begin{align}
\mathbb{P}\Big(\bigcup_{t=0}^{K-1}\mathcal{E}_t^c\Big)
\le \sum_{t=0}^{K-1}\mathbb{P}(\mathcal{E}_t^c)
= \sum_{t=0}^{K-1}\mathbb{E}\big[\mathbf{1}_{\mathcal{E}_t^c}\big]
= \sum_{t=0}^{K-1}\mathbb{E}\Big[\mathbb{E}\big[\mathbf{1}_{\mathcal{E}_t^c}\,\big|\,\mathcal{F}_t\big]\Big]
= \sum_{t=0}^{K-1}\mathbb{E}\big[\mathbb{P}(\mathcal{E}_t^c\mid\mathcal{F}_t)\big]
\le K\bar\delta , \label{eq:pf-step3}
\end{align}
 
\smallskip\noindent\emph{Step 4 (conclusion).}
On $\mathcal{E}_t$, inequality~\eqref{eq:pf-step2} and the definition~\eqref{eq:cbar} of $\bar c$ give
\begin{equation*}
V_1(\btheta_{t+1}) \le (1-\alpha h_t)V_1(\btheta_t)
+\alpha h_t(\zeta_1+\epsilon)
+\frac{h_t(\sqrt{d}\,\tilde\zeta_1+\epsilon)^2}{2(\beta-L_1h_{\max})}
=(1-\alpha h_t)V_1(\btheta_t)+\alpha h_t\,\bar c ,
\end{equation*}
or, equivalently,
$V_1(\btheta_{t+1})-\bar c\le(1-\alpha h_t)\big(V_1(\btheta_t)-\bar c\big)$.
On $\bigcap_{t=0}^{K-1}\mathcal{E}_t$, which by~\eqref{eq:pf-step3} has probability at least $1-K\bar\delta$, this holds for all $t\in0:K-1$.
Since $0<1-\alpha h_t<1$ (because $0<h_t<1/\alpha$), iterating it yields
$V_1(\btheta_t)-\bar c\le\Pi_t\big(V_1(\btheta_0)-\bar c\big)$ for all $t\in 0:K$, which is~\eqref{eq:multistep}.
Item (i) follows because $\Pi_tV_1(\btheta_0)\le0$ and $1-\Pi_t\le1$ when $V_1(\btheta_0)\le0$, and item (ii) follows from $1-\Pi_t\le1$.
\end{proof}

\section{Proof of Proposition~\ref{prop:conv}}\label{app:proof-proposition-conv}

Throughout this section, we let 
$\pos{x}=\max\{x,0\}$, $\Rc_t:=\Rc(\btheta_t)$,
$\hat{\Rc}_t:=\hat{\Rc}(\btheta_t)$, $E_t:=\| \hat{\Rc}_t-\Rc_t \|$, and $\E_t[\cdot]:=\E[\cdot\mid\mathcal F_t]$ in order to simplify notation. We also drop the dependency in $\btheta$ wherever it is clear.
Given $\bp\in\real^d$ and a compact set $A\subset\real^d$, $\text{Proj}_A(\bp)$ denotes the projection of $\bp$ onto $A$.

We introduce a few technical lemmas which are crucial for the proof of Proposition~\ref{prop:conv}.

\begin{lemma}\longthmtitle{Projection interpretation of RSGF}\label{lem:ball}
For $v\in\R$, $\bp,\bq\in\R^d$, the set
$\{\bxi\in\real^d:\alpha v+\bq^\top\bxi+\frac\beta2\|\bxi\|^2\le0\}$ is nonempty iff
$\|\bq\|^2/\beta^2-2\alpha v/\beta\ge0$, in which case it equals the closed ball
$\mathbb B(-\bq/\beta,\,r)$ with $r = \sqrt{ \|\bq\|^2/\beta^2-2\alpha v/\beta }$. Hence, whenever~\eqref{eq:R-definition} is feasible we have
\begin{align*}
    \Rc(\btheta)=P_{\mathbb B(-\nabla V_1(\btheta)/\beta,\,r(\btheta))}(-\nabla V_0(\btheta)).
\end{align*}
\end{lemma}
\begin{proof}
Completing the square,
$\alpha v+\bq^\top\bxi+\frac\beta2\| \bxi \|^2=\frac\beta2\big(\|\bxi+\bq/\beta\|^2-r^2\big)$,
and the objective function $\frac{1}{2}\norm{\bxi + \nabla V_0(\btheta)}^2$ is minimized over the closed convex set $\mathbb{B}(-q/\beta,r)$ by the projection of $-\nabla V_0(\btheta)$ onto this set $\mathbb{B}(-q/\beta,r)$.
\end{proof}

\begin{lemma}\longthmtitle{Multiplier bound}\label{lem:mult}
Suppose that $\Theta\subset\R^d$ is compact and that assumption~(iii) of Proposition~\ref{prop:conv} holds, and let $\bar u$ be as defined in~\eqref{eq:consts}.
Let $\btheta\in\Theta$ and let $u(\btheta)$ be the (unique) multiplier of~\eqref{eq:R-definition}. Then
$u(\btheta)\le\bar u$. 
\end{lemma}
\begin{proof}
The center $\bar\bxi=-\nabla V_1/\beta$ satisfies
\begin{align*}
    \alpha V_1(\theta) + \nabla V_1(\theta)^\top \bar{\xi} + \frac{\beta}{2}\norm{\bar{\xi}}^2 = -\tfrac\beta2 D(\btheta)\le-\tfrac\beta2\rho^2<0.
\end{align*}
Hence, Slater's condition holds and $\Rc(\btheta)$ admits a KKT
multiplier $u\ge0$. Since $u\ge0$, the Lagrangian
\begin{align*}
    L(\bxi,u)=\tfrac12\|\bxi+\nabla V_0\|^2+u\Big(
    \alpha V_1 + \nabla V_1^\top \xi + \frac{\beta}{2}\norm{ \xi }^2
    \Big)
\end{align*}
is convex in $\bxi$, and stationarity gives
$\nabla_\bxi L(\Rc,u)=0$, so $\Rc$ minimizes $L(\cdot,u)$. Together with complementary slackness
of~\eqref{eq:R-definition}
this yields
$\tfrac12\|\Rc+\nabla V_0\|^2=L(\Rc,u)\le L(\bar\bxi,u)\le\tfrac12\|\bar\bxi+\nabla V_0\|^2-u\tfrac\beta2\rho^2$,
and therefore $u\le\|\nabla V_0-\nabla V_1/\beta\|^2/(\beta\rho^2)\le\bar u$.


\end{proof}
 
\begin{lemma}\longthmtitle{RSGF perturbation bound}\label{lem:pert}
Suppose that $\Theta\subset\R^d$ is compact and that assumption~(iii) of Proposition~\ref{prop:conv} holds, and let $c_1,c_V$ be as defined in~\eqref{eq:consts} and $\varepsilon_N$ as defined in~\eqref{eq:epsN}.
Let $\btheta\in\Theta$, $e_V:=\Vh(\btheta)-V_1(\btheta)$, $\be_j:=\gh j(\btheta)-\nabla V_j(\btheta)$. Then,
almost surely,
\begin{equation}\label{eq:pert}
  \|\hat{\Rc}(\btheta)-\Rc(\btheta)\|\le\|\be_0\|+c_1\|\be_1\|+c_V|e_V|.
\end{equation}
If, in addition, $\theta_t\in\Theta$ almost surely for some $t\in\mathbb{N}$,
then $\E_t[E_t^2]\le\varepsilon_N^2$, $\E_t[E_t]\le\varepsilon_N$ and
$\| \Rc(\theta_t)-\E_t[\hat{\Rc}(\theta_t)]\|\le\varepsilon_N$.
%
\end{lemma}
\begin{proof}
Throughout, we let $P(\by;\bc,s):=\text{Proj}_{\mathbb B(\bc,s)}(\by)$. We also write $\hat s(\btheta):=\sqrt{\pos{\hat D(\btheta)}}$.
By Lemma~\ref{lem:ball} and since $\hat{\Rc}(\theta) = -\widehat{\nabla V}_1(\theta) / \beta$ by convention whenever~\eqref{eq:R-hat-definition} is infeasible, we have
\begin{equation}\label{eq:RRhproj}
  \Rc(\btheta)=P\big(-\nabla V_0(\btheta);\,-\nabla V_1(\btheta)/\beta,\,\sqrt{D(\btheta)}\big),\qquad
  \hat{\Rc}(\theta)=P\big(-\gh0(\theta);\,-\gh1(\theta)/\beta,\,\hat s(\theta)\big),
\end{equation}
where $\Rc(\theta)$ is well defined since $r(\theta)\ge\rho>0$ for all $\theta\in\Theta$ by assumption. We split the proof in three steps.
 
\smallskip\noindent\emph{Step 1 (Lipschitz dependence of ball projections).}
Let us show that for all $\by,\by',\bc,\bc'\in\R^d$ and $s,s'\ge0$,
\begin{equation}\label{eq:projlip}
  \|P(\by;\bc,s)-P(\by';\bc',s')\|\le\| \by-\by'\|+\| \bc-\bc'\|+|s-s'| .
\end{equation}
To prove it, we use three standard facts. First, for any nonempty, closed, convex $C\subseteq\R^d$, both
$\text{Proj}_C$ and $I-\text{Proj}_C$ are (firmly) nonexpansive
\cite[Prop.~4.16 and~4.4]{HHB-PLC:17}. Second, we have
$P(\by;\bc,s)=\bc+P(\by-\bc;0,s)$~\cite[Prop.~29.1]{HHB-PLC:17}. Third,
$P(0;0,s)=0$ and $P(\bz;0,s)=\min\{s,\| \bz \|\}\, \bz/ \|\bz\|$ for $\bz\neq0$. 
By the triangle inequality, 
\begin{align*}
    \norm{ P(\by;\bc,s)-P(\by^\prime;\bc^\prime,s^\prime) } &\leq 
    \|P(\by;\bc,s)-P(\by';\bc,s)\| + 
    \|P(\by';\bc,s)-P(\by';\bc',s)\|
    \\
    &\quad +\|P(\by';\bc',s)-P(\by';\bc',s')\|
\end{align*}
Now, by the non-expansiveness, we have 
\begin{align*}
    \norm{ P(\by;\bc,s)-P(\by^\prime;\bc,s) } \leq \norm{\by - \by^\prime}.
\end{align*}

With $Q:=I-\text{Proj}_{\mathbb B(0,s)}$, the translation property gives
$P(\by';\bc,s)=\by'-Q(\by'-\bc)$, 
so the (firm) nonexpansiveness of $Q$ yields
\begin{align*}
    \|P(\by^\prime;\bc,s)-P(\by';\bc^\prime,s)\| = 
    \| Q(\by^\prime-\bc^\prime)-Q(\by^\prime-\bc) \| \leq \norm{\bc-\bc^\prime}.
\end{align*}
Finally,
\begin{align*}
    \|P(\by';\bc',s)-P(\by';\bc',s')\| &= \| P(\by'-\bc';0,s)-P(\by'-\bc';0,s') \| \\
    &=| \min\{s,\norm{\by^\prime-\bc^\prime} \}\frac{\by^\prime-\bc^\prime}{\norm{\by^\prime-\bc^\prime}} - \min\{s^\prime,\norm{\by^\prime-\bc^\prime} \}\frac{\by^\prime-\bc^\prime}{\norm{\by^\prime-\bc^\prime}} | \\
    &= | \min\{s,\norm{\by^\prime-\bc^\prime} \} - \min\{s^\prime,\norm{\by^\prime-\bc^\prime} \} | \leq |s-s^\prime|,
\end{align*}
from where~\eqref{eq:projlip} follows. 
Now, applying~\eqref{eq:projlip} to~\eqref{eq:RRhproj} with $\by=-\gh0$, $\bc=-\gh1/\beta$,
$s=\hat s$ and $\by'=-\nabla V_0$, $\bc'=-\nabla V_1/\beta$, $s'=\sqrt{D}$ (all evaluated at $\theta$),
\begin{equation}\label{eq:step2}
  \| \hat{\Rc}-\Rc\| \le \|\be_0\|+\frac{\|\be_1\|}{\beta}+|\hat s-\sqrt{D}|.
\end{equation}
 
\smallskip\noindent\emph{Step 2 (bounding the radius error).}
We first note that for $a\in\R$ and $b>0$,
\begin{equation}\label{eq:sqrt}
  \big|\sqrt{\pos a}-\sqrt b\big|\le\frac{|a-b|}{\sqrt b}.
\end{equation}
If $a\ge0$, this follows from $|\sqrt a-\sqrt b|=|a-b|/(\sqrt a+\sqrt b)$ and
$\sqrt a+\sqrt b\ge\sqrt b$. If $a<0$, then $|\sqrt{\pos a}-\sqrt b|=\sqrt b=b/\sqrt b\le(b-a)/\sqrt b$.
Applying~\eqref{eq:sqrt} with $a=\hat D(\btheta)$, $b=D(\btheta)\ge\rho^2>0$ gives
$|\hat s-\sqrt{D}|\le|\hat D-D|/\rho$. By~\eqref{eq:radius},
\[
  |\hat D-D|\le\frac{\big|\|\gh1\|^2-\|\nabla V_1\|^2\big|}{\beta^2}+\frac{2\alpha|e_V|}{\beta}.
\]
Moreover,
$\big|\|\gh1\|^2-\|\nabla V_1\|^2\big|=|(\gh1-\nabla V_1)^\top(\gh1+\nabla V_1)|
\le\|\be_1\|\,(\|\gh1\|+\|\nabla V_1\|)\le\|\be_1\|(\sqrt d\,\tilde B_1+G_1)$, where we used
the Cauchy--Schwarz inequality, $\|\nabla V_1\|\le G_1$,
and $\|\gh1\|\le\sqrt d\,\tilde B_1$ a.s.
Therefore
\[
  |\hat s-\sqrt{D}|\le\frac{\sqrt d\,\tilde B_1+G_1}{\beta^2\rho}\|e_1\|+\frac{2\alpha}{\beta\rho}|e_V|,
\]
and substituting into~\eqref{eq:step2} yields~\eqref{eq:pert} with $c_1,c_V$ as
in~\eqref{eq:consts}.
 
\smallskip\noindent\emph{Step 3 (moment bounds).}
Let $\theta=\theta_t$, which is $\mathcal F_t$-measurable. Squaring~\eqref{eq:pert} and using
$(a+b+c)^2\le3(a^2+b^2+c^2)$ for all $a, b, c > 0$,
\begin{equation}\label{eq:Esq}
  \E_t[E_t^2]\le3\big(\E_t\|e_0\|^2+c_1^2\,\E_t\|e_1\|^2+c_V^2\,\E_t[e_V^2]\big).
\end{equation}
Fix $j\in\{0,1\}$ and write $\gh j=\frac1N\sum_{n=1}^N\tilde G^{\theta_t}_{j,n}$. Now, since $\E_t[\gh j-\E_t[\gh j]]=0$, we have
\begin{align*}
    \E_t\|e_j\|^2=\|\E_t[\gh j]-\nabla V_j(\theta_t)\|^2+\E_t\|\gh j-\E_t[\gh j]\|^2.
\end{align*}
For the first term, by~\eqref{eq:bounded-bias} each coordinate of $\E_t[\gh j]-\nabla V_j(\theta_t)$ is at most
$\tilde\zeta_j$ in absolute value, so the term is at most $d\tilde\zeta_j^2$. For the second term,
$\gh j-\E_t[\gh j]=\frac1N\sum_n(\tilde G_{j,n}^{\theta_t}-\E_t\tilde G_{j,n}^{\theta_t})$. By
assumption, the $\tilde G_{j,n}^{\theta_t}$ are conditionally independent given
$\mathcal F_t$, so
$\E_t\langle\tilde G_{j,n}^{\theta_t}-\E_t\tilde G_{j,n}^{\theta_t},\tilde G_{j,m}^{\theta_t}-\E_t\tilde G_{j,m}^{\theta_t}\rangle=0$ for $n\ne m$, and
\begin{align*}
    &\E_t\|\gh j-\E_t[\gh j]\|^2=\frac1{N^2}\sum_{n=1}^N\E_t\|\tilde G_{j,n}^{\theta_t}-\E_t\tilde G_{j,n}^{\theta_t}\|^2
    =\frac{1}{N^2}\sum_{n=1}^N \Big( \E_t \norm{\tilde{G}_{j,n}^{\theta_t}}^2 - \norm{ \E_t \tilde{G}_{j,n}^{\theta_t} }^2 \Big) \\
    &\le\frac1{N^2}\sum_{n=1}^N\E_t\|\tilde G_{j,n}^{\theta_t} \|^2\le\frac{d\tilde B_j^2}{N},
\end{align*}

where we used $\|\tilde G_{j,n}^{\theta_t}\|^2\le d\tilde B_j^2$ a.s.\ by~\eqref{eq:boundedness-estimates}. Hence
$\E_t\|e_j\|^2\le d(\tilde\zeta_j^2+\tilde B_j^2/N)$. The same argument applied to the
scalar samples $G^{\theta_t}_{1,n}$, using $|\E_t[\Vh]-V_1(\theta_t)|\le\zeta_1$ and
$|G^{\theta_t}_{1,n}|\le B_1$, gives $\E_t[e_V^2]\le\zeta_1^2+B_1^2/N$. Substituting
into~\eqref{eq:Esq} gives $\E_t[E_t^2]\le\varepsilon_N^2$. By Jensen's inequality,
$\E_t[E_t]\le(\E_t[E_t^2])^{1/2}\le\varepsilon_N$. Finally, by Jensen's inequality we have
$\| \Rc(\theta_t)-\E_t[\hat{\Rc}(\theta_t)]\|=\|\E_t[ \hat{\Rc}(\theta_t)-\Rc(\theta_t)]\|\le\E_t[E_t]\le\varepsilon_N$.
 

\end{proof}
 
\begin{lemma}\longthmtitle{$\Phi_{\kappa}$ function descent}\label{lem:descent}
Suppose that assumptions~(i)--(iii) of Proposition~\ref{prop:conv} hold, fix $\kappa\ge\bar u$, and let $\Phi_\kappa$, $G_\Phi$ and $L_\Phi$ be as defined in Proposition~\ref{prop:conv}.
If $h_t\le1/\alpha$ then, 
\begin{equation}\label{eq:descent}
  \Phi_\kappa(\theta_{t+1})\le\Phi_\kappa(\theta_t)-h_t\| \Rc_t\|^2+h_tG_\Phi E_t
  +\frac{L_\Phi h_t^2}{2}\| \hat{\Rc}_t \|^2 \quad a.s.
\end{equation}
If moreover $h_t\le1/(2L_\Phi)$, then
$\Phi_\kappa(\theta_{t+1})\le\Phi_\kappa(\theta_t)-\frac{h_t}{2}\| \Rc_t\|^2+h_tG_\Phi E_t+L_\Phi h_t^2E_t^2$.
\end{lemma}
\begin{proof}
Throughout the proof, all gradients are evaluated at $\theta_t$ unless stated otherwise. We let $\Delta_t:=\theta_{t+1}-\theta_t=h_t\hat{\Rc}_t$, and $u_t:=u(\theta_t)$ be the multiplier of~\eqref{eq:R-definition} at $\theta_t$, which by Lemma~\ref{lem:mult} satisfies $0\le u_t\le\bar u$. No expectation is taken in this proof: all inequalities hold for every realization with $\theta_t,\theta_{t+1}\in\Theta$ (by assumption, this means that all inequalities hold almost surely).

Since $\theta_t,\theta_{t+1}\in\Theta$, $\Theta$ is convex and $\nabla V_j$ is $L_j$-Lipschitz on $\Theta$,~\cite[Lemma 1.2.3]{YN:18} gives, for $j\in\{0,1\}$,
\begin{align*}
  V_j(\theta_{t+1})
  &\le V_j(\theta_t)+\nabla V_j(\theta_t)^\top\Delta_t+\tfrac{L_j}{2}\|\Delta_t\|^2 \\
  &= V_j(\theta_t)+h_t\nabla V_j(\theta_t)^\top\Rc_t+h_t\nabla V_j(\theta_t)^\top(\hat{\Rc}_t-\Rc_t)+\tfrac{L_j}{2}\|\Delta_t\|^2 .
\end{align*}
By the Cauchy--Schwarz inequality and $\|\nabla V_j(\theta_t)\|\le G_j$, we have $h_t\nabla V_j(\theta_t)^\top(\hat{\Rc}_t-\Rc_t)\le h_tG_jE_t$, so
\begin{equation}\label{eq:descent-Vj}
  V_j(\theta_{t+1})\le V_j(\theta_t)+h_t\nabla V_j(\theta_t)^\top\Rc_t+h_tG_jE_t+\tfrac{L_j}{2}\|\Delta_t\|^2,\qquad j\in\{0,1\}.
\end{equation}
The map $\pos{\cdot}$ is nondecreasing, subadditive ($\pos{a+b}\le\pos a+\pos b$), and satisfies $\pos c=c$ for $c\ge0$. Applying these properties to~\eqref{eq:descent-Vj} with $j=1$ (the last two terms there are nonnegative) gives
\begin{equation*}
  \pos{V_1(\theta_{t+1})}\le\pos{V_1(\theta_t)+h_t\nabla V_1(\theta_t)^\top \Rc_t}+h_tG_1E_t+\tfrac{L_1}{2}\|\Delta_t\|^2 .
\end{equation*}
Multiplying this inequality by $\kappa\ge0$ and adding~\eqref{eq:descent-Vj} with $j=0$ yields
\begin{equation}\label{eq:descent-Phi}
  \Phi_\kappa(\theta_{t+1})\le\Psi_t+h_tG_\Phi E_t+\tfrac{L_\Phi h_t^2}{2}\|\hat{\Rc}_t\|^2,
\end{equation}
where
\[
  \Psi_t:=V_0(\theta_t)+h_t\nabla V_0^\top\Rc_t+\kappa\pos{V_1(\theta_t)+h_t\nabla V_1^\top\Rc_t},
\]
and we used $G_\Phi=G_0+\kappa G_1$, $L_\Phi=L_0+\kappa L_1$ and $\|\Delta_t\|^2=h_t^2\|\hat{\Rc}_t\|^2$. The result follows if we show that
\begin{equation}\label{eq:descent-key}
  \Psi_t\le\Phi_\kappa(\theta_t)-h_t\|\Rc_t\|^2 .
\end{equation}

To see this, note that by~\eqref{eq:R-definition}:
\begin{equation}\label{eq:descent-V1-lin}
  V_1(\theta_t)+h_t\nabla V_1^\top\Rc_t\le(1-\alpha h_t)V_1(\theta_t)-\tfrac{\beta h_t}{2}\|\Rc_t\|^2\le(1-\alpha h_t)V_1(\theta_t).
\end{equation}
On the other hand, as shown in~\cite[Proposition IV.5 (i)]{PM-AM-JC:25-arxiv} we have
\begin{equation}\label{eq:descent-V0-lin}
  h_t\nabla V_0^\top\Rc_t\le-h_t\|\Rc_t\|^2+u_t\,\alpha h_t\,V_1(\theta_t).
\end{equation}

Now we distinguish two cases:

\smallskip\noindent\emph{Case $V_1(\theta_t)\le0$.} Combining~\eqref{eq:descent-V1-lin},~\eqref{eq:descent-V0-lin}, $V_1(\theta_t) \le 0$ and the definition of $[\cdot]$ we get
\begin{align*}
    \Psi_t \leq V_0(\theta_t) - h_t \norm{\Rc_t}^2 \leq \Phi_{\kappa}(\theta_t) - h_t \norm{\Rc_t}^2,
\end{align*}
where in the last inequality we used that $[V_1(\theta_t)]_{+}\geq0$. Hence,~\eqref{eq:descent-key} holds.

\smallskip\noindent\emph{Case $V_1(\theta_t)>0$.}
Combining~\eqref{eq:descent-V1-lin} and~\eqref{eq:descent-V0-lin} we get
\begin{align*}
  \Psi_t
  &\le V_0(\theta_t)-h_t\|\Rc_t\|^2+\bar u\alpha h_tV_1(\theta_t)+\kappa V_1(\theta_t)-\kappa\alpha h_tV_1(\theta_t) \\
  &=\Phi_\kappa(\theta_t)-h_t\|\Rc_t\|^2-(\kappa-\bar u)\,\alpha h_t\,V_1(\theta_t)
  \le\Phi_\kappa(\theta_t)-h_t\|\Rc_t\|^2,
\end{align*}
where the last inequality uses $\kappa\ge\bar u$.
\end{proof}
 
\begin{proof}[Proof of Proposition~\ref{prop:conv}]
\ref{it:non-asymptotic-bound} Summing the second inequality of Lemma~\ref{lem:descent} over $t=0,\dots,K-1$ and
using $\Phi_\kappa(\theta_K)\ge\min_\Theta V_0$,
\begin{equation}\label{eq:pathsum}
  \frac12\sum_{t=0}^{K-1}h_t\|\Rc_t\|^2\le\Delta_0+G_\Phi\sum_{t=0}^{K-1}h_tE_t+L_\Phi\sum_{t=0}^{K-1}h_t^2E_t^2,
\end{equation}
for every realization. Taking expectations and using $\E[E_t]\le\varepsilon_N$,
$\E[E_t^2]\le\varepsilon_N^2$ (Lemma~\ref{lem:pert} and the tower property) and dividing by
$H_K/2$ gives~\eqref{eq:main}. For constant $h$, $2L_\Phi h\le1$ gives~\eqref{eq:const}.
 
\ref{it:almost-sure-bound} 
By Lemma~\ref{lem:ball} (and the convention for infeasible~\eqref{eq:R-hat-definition}), $\hat{\Rc}_t$ lies in the closed ball with
center $-\widehat{\nabla V}_1(\theta_t)/\beta$ and radius $\sqrt{[\hat D(\theta_t)]_+}$. Since
$\|\widehat{\nabla V}_1(\theta_t)\|\le\sqrt d\,\tilde B_1$ and $|\hat V_1(\theta_t)|\le B_1$ a.s. by~\eqref{eq:boundedness-estimates},
so $\sqrt{[\hat{D}(\theta_t)]_{+}} \leq \sqrt{d \tilde{B}_1^2 / \beta^2 + 2\alpha B_1 / \beta} \leq \sqrt{d}\tilde{B}_1 / \beta + \sqrt{2\alpha B_1 / \beta}$, which means that
$\|\hat{\Rc}_t\|\le 2\sqrt d\,\tilde B_1/\beta+\sqrt{2\alpha B_1/\beta}$ a.s. Moreover, $\Rc$ is continuous
on the compact set $\Theta$ (by Lemma~\ref{lem:pert}) and $\theta_t\in\Theta$ a.s., so $\|\Rc_t\|\le\max_{\Theta}\|\Rc\|<\infty$.
Hence $E_t\le\|\hat{\Rc}_t\|+\|\Rc_t\|\le\bar E$ a.s. for a deterministic constant $\bar E$.
Hence, $D_t:=E_t-\E_t[E_t]$ and $D_t':=E_t^2-\E_t[E_t^2]$ are bounded martingale differences adapted to $\mathcal F_{t+1}$. With
$H_{t+1}=\sum_{k\le t}h_k$, we have $(h_t/H_{t+1})^2\le\bar h\,(1/H_t-1/H_{t+1})$ for $t\ge1$,
so $\sum_t(h_t/H_{t+1})^2<\infty$ and therefore
the martingales 
$\sum_{k \leq t}(h_k/H_{k+1})D_k$,
$\sum_{k \leq t}(h_k/H_{k+1})D'_k$ are bounded in $L^2$, hence converge a.s by~\cite[Theorem 4.4.6]{RD:10-5th-edition}.
Since $H_K\to\infty$,
Kronecker's lemma~\cite[Theorem 2.5.9]{RD:10-5th-edition} yields
\begin{align*}
    \frac1{H_K}\sum_{t<K}h_tD_t\to0, \quad \frac1{H_K}\sum_{t<K}h_tD'_t\to0, \quad a.s.,
\end{align*}
so
$\limsup_K\frac1{H_K}\sum_{t<K}h_tE_t\le\varepsilon_N$ and
$\limsup_K\frac1{H_K}\sum_{t<K}h_tE_t^2\le\varepsilon_N^2$ a.s. 
Dividing~\eqref{eq:pathsum} by
$H_K/2$, using $L_\Phi h_t^2E_t^2\le\frac12h_tE_t^2$ and letting $K\to\infty$
gives~\eqref{eq:as}. 

For the $\liminf$ statement, fix a realization for which~\eqref{eq:as} holds, and let
$a_t:=\|\Rc(\theta_t)\|^2\ge0$ and $c:=2G_\Phi\varepsilon_N+\varepsilon_N^2$.
Suppose, by contradiction, that $\liminf_{t\to\infty}a_t>c$. Then there exist $m>c$ and $T\in\mathbb N$
such that $a_t\ge m$ for all $t\ge T$. Since $a_t\ge0$ and $h_t>0$, for every $K>T$,
\[
  \frac{1}{H_K}\sum_{t=0}^{K-1}h_ta_t\;\ge\;\frac{1}{H_K}\sum_{t=T}^{K-1}h_ta_t
  \;\ge\; m\,\frac{H_K-H_T}{H_K}.
\]
Since $H_T$ is fixed and $H_K\to\infty$, the right-hand side converges to $m$ as $K\to\infty$, so
$\limsup_{K\to\infty}\frac{1}{H_K}\sum_{t=0}^{K-1}h_ta_t\ge m>c$, which contradicts~\eqref{eq:as}.
Hence $\liminf_{t\to\infty}\|\Rc(\theta_t)\|^2\le c$ for every realization for which~\eqref{eq:as} holds,
i.e., almost surely.
\end{proof}

\renewcommand{\floatpagefraction}{0.8}
\section{Experiments reward definitions}\label{sec:experiments-reward-definitions}

Table~\ref{tab:rewards} lists the reward terms for all experiments. Each term contributes $w \cdot r$ to the total reward, where $r$ is the expression shown and $w$ its weight. All expressions are non-negative; penalties carry negative weights.

For the humanoid squatting and crawling, terms follow BeyondMimic~\citep{QL-TET-XH-YG-GT-KS-KCL:25} as implemented in mjlab~\citep{KZ-QL-BY-LLL-KS-PA:26}:
$p_r,q_r$ root position/orientation; $p_i,q_i,v_i,\omega_i$ position, orientation, linear and angular velocity of tracked body $i$;
$d_q$ geodesic quaternion distance; $c_{\mathrm{joint}}$ soft joint-limit violation; $F_{i,b}$ contact force on body $h$.
$\Delta x = x^{\mathrm{ref}}-x$; $\overline{(\cdot)}$ is the mean over the $N_b$ tracked bodies.

For the humanoid walking, generic terms follow~\citep{NR-DH-PR-MH:22}; the gait term follows~\citep{JS-YG-AF-JH:21}.
$v_x^\star$ commanded forward speed; $e_y,e_\psi$ lateral and heading drift; $s_f,s_f^\star$ stride length and target;
$s_\ell$ lateral stride offset; $d_L,d_R$ left/right leg deviation; $h_i,h^\star$ swing-foot height and target;
$c_i$ contact of foot $i$; $\mathrm{td},i$ touchdown of foot $i$; $g_{b,xy}$ projected gravity in base frame; $z_b$ base height;
$\mathcal{G}_\sigma(e)=e^{-e^2/(2\sigma^2)}$; $a\!\to\!b$ denotes a curriculum.

For the Crazyflie experiment, $p,v$ denote drone position and velocity, $p^{\mathrm{ref}}$ the hover target, $d=\|p-p^{\mathrm{ref}}\|$, $v_b,\omega_b$ body-frame linear and angular velocities, and $a_t,\ddot p_t$ action and acceleration at step $t$.
$V$ denotes a quadratic CLF
on position and attitude errors and their rates~\citep{li2026clf}; $\bar\lambda=\lambda_{\max}(P)$, and $\Delta_{\max}$ clips the decrease condition $\dot V\le-0.02\,V$.
CAPS~\citep{mysore2021regularizing} uses $\mu_t=\mu(s_t)$ policy mean action, $\epsilon$ small random state perturbation, $\operatorname{sg}[\cdot]$ stop-gradient.
We also denote by $p_i$, $v_i$, $r_i$ the center, velocity, and radius of obstacle $i$; drone radius $r=0.05$\,m safety margin $m=0.1$\,m, $c_{\mathrm{obs}}=\min_i\|p_i-p\|-(\rho_i+r+m)$, and $c_{\mathrm{arena}}$ is the clearance to the arena boundary minus $m$.
The safety reward is $R_1=1-\frac{\delta}{T+1}$ outside the safe set $\bar{\Cc}=\{\min(c_{\mathrm{obs}},c_{\mathrm{arena}})\ge0\}$ (Table~\ref{tab:experiment-params}); episodes terminate only on contact.
$\hat c_{\mathrm{obs}}=\max\{0,\min_i\min_{t\in[0,\tau]}\|(p_i-p)+t(v_i-v)\|-(\rho_i+r+m)\}$, with $\tau=0.5$\,s, is the clearance predicted at constant relative velocity, so $\tilde h\ge 0$ on $\bar{\Cc}$.

\begin{table}[!htbp]
\centering\footnotesize
\setlength{\tabcolsep}{3.5pt}
\begin{tabular}{@{}llr@{}}
\multicolumn{3}{@{}l}{\textbf{(a) Humanoid squatting and crawling (motion tracking)}} \\
\toprule
\textbf{Term} & \textbf{Expression} & \textbf{Weight} \\
\midrule
\multicolumn{3}{@{}l}{\emph{Task reward $R_0$}} \\
Root position & $e^{-\|\Delta p_r\|^2/0.3^2}$ & $0.5$ \\
Root orientation & $e^{-d_q(q_r^{\mathrm{ref}},q_r)^2/0.4^2}$ & $0.5$ \\
Body position & $e^{-\overline{\|\Delta p_i\|^2}/0.3^2}$ & $1.0$ \\
Body orientation & $e^{-\overline{d_q(q_i^{\mathrm{ref}},q_i)^2}/0.4^2}$ & $1.0$ \\
Body lin.\ velocity & $e^{-\overline{\|\Delta v_i\|^2}/1.0^2}$ & $1.0$ \\
Body ang.\ velocity & $e^{-\overline{\|\Delta\omega_i\|^2}/3.14^2}$ & $1.0$ \\
Action rate & $\|a_t-a_{t-1}\|^2$ & $-0.2$ \\
Joint limits & $c_{\mathrm{joint}}$ & $-10$ \\
Self-collision & $\sum_{b=1}^{B}\mathbb{I}\big[\max_i\|F_{i,b}\|>10\big]$ & $-10$ \\
\midrule
\multicolumn{3}{@{}l}{\emph{Safety reward $R_1$}} \\
Base / head height & $R_1$ in~\eqref{eq:R1-practical-choice-probabilistic-safety}, $\delta=0.1$ & -- \\
\bottomrule
\end{tabular}

\medskip
\resizebox{\linewidth}{!}{%
\begin{tabular}{@{}llr@{\hspace{10pt}}llr@{}}
\multicolumn{6}{@{}l}{\textbf{(b) Humanoid walking}} \\
\toprule
\textbf{Term} & \textbf{Expression} & \textbf{Weight} & \textbf{Term} & \textbf{Expression} & \textbf{Weight} \\
\midrule
\multicolumn{3}{@{}l}{\emph{Task}} & \multicolumn{3}{l}{\emph{Posture / stability}} \\
Forward velocity & $\mathcal{G}_{\sigma_v}(v_x-v_x^\star)$ & $1\!\to\!3$ &
Arm posture & $\operatorname{mean}\big((q-q^\star)^2/\sigma^2\big)$ & $-1$ \\
Straight line & $\mathcal{G}_{\sigma_y}(e_y)\,\mathcal{G}_{\sigma_\psi}(e_\psi)$ & $1.5\!\to\!3$ &
Leg posture & $\operatorname{mean}\big((q-q^\star)^2/\sigma^2\big)$ & $-0.7$ \\
\multicolumn{3}{@{}l}{\emph{Gait}} &
Lateral motion & $v_{b,y}^2+v_{b,z}^2$ & $-0.5$ \\
Foot gait & phase-dependent contacts & $1$ &
Orientation rate & $\omega_x^2+\omega_y^2$ & $-1$ \\
Stride length & $\mathcal{G}_{\sigma_s}(s_f-s_f^\star)\,\mathcal{G}_{\sigma_\ell}(s_\ell)$ & $1$ &
Upright & $\mathcal{G}_{0.15}(\|g_{b,xy}\|)$ & $1.8$ \\
Leg symmetry & $e^{-\operatorname{mean}(d_L+d_R)^2/\sigma_{\mathrm{sym}}^2}$ & $3$ &
Base height & $\mathcal{G}_{0.05}(z_b-0.793)$ & $1$ \\
Foot clearance & $\sum_i |h_i-h^\star|\,\|\dot p_{i,xy}\|$ & $-2$ &
\multicolumn{3}{l}{\emph{Regularization}} \\
Foot slip & $\sum_i (\|\dot p_{i,xy}\|^2+0.25\,\omega_{i,z}^2)\,1_{c_i}$ & $-1$ &
Alive & $1$ & $0.05$ \\
Soft landing & $\sum_i \|F_i\|\,1_{\mathrm{td},i}$ & $-10^{-3}$ &
Action & $\|a_t\|^2$ & $-0.1$ \\
& & & Action rate & $\|a_t-a_{t-1}\|^2$ & $-0.15$ \\
& & & Joint acceleration & $\|\ddot q\|^2$ & $-10^{-6}$ \\
& & & Joint limits & soft joint-limit violation & $-0.5$ \\
\midrule
\multicolumn{6}{@{}l}{\emph{Safety reward $R_1$}} \\
Height / attitude / CoP & $R_1$ in~\eqref{eq:R1-practical-choice-probabilistic-safety}, $\delta=0.1$ & -- & & & \\
\bottomrule
\end{tabular}}

\medskip
\begin{tabular}{@{}llr@{}}
\multicolumn{3}{@{}l}{\textbf{(c) Crazyflie obstacle avoidance}} \\
\toprule
\textbf{Term} & \textbf{Expression} & \textbf{Weight} \\
\midrule
\multicolumn{3}{@{}l}{\emph{Task reward $R_0$}} \\
Settle & $\big(1-\tanh\tfrac{\|p-p^{\mathrm{ref}}\|}{0.2}\big)\big(1-\tanh\tfrac{\|v\|}{0.3}\big)$ & $20$ \\
Target distance & $10-d$ & $0.5$ \\
CLF & $e^{-V/(0.7^2\bar\lambda)}$ & $10$ \\
CLF decrease & $\operatorname{clip}\big([\dot V+0.02\,V]_+/\Delta_{\max},\,0,\,1\big)$ & $-5$ \\
Lin.\ / ang.\ velocity & $\|v_b\|^2$ \,/\, $\|\omega_b\|^2$ & $-0.01$ / $-0.02$ \\
Action rate & $\|a_t-a_{t-1}\|^2$ & $-0.01$ \\
Jerk & $\|\ddot{p}_t-\ddot{p}_{t-1}\|^2$ & $-10^{-4}$ \\
\midrule
\multicolumn{3}{@{}l}{\emph{Policy-loss regularization (CAPS)}} \\
Temporal / spatial & $\|\mu_t-\mu_{t-1}\|^2$ \,/\, $\|\mu(s_t+\epsilon)-\operatorname{sg}[\mu(s_t)]\|^2$ & $1$ / $1$ \\
\midrule
\multicolumn{3}{@{}l}{\emph{Safety reward $R_1$}} \\
Hazard proximity & $R_1$ in~\eqref{eq:R1-practical-choice-probabilistic-safety} with $\delta=0.05$ and $\tilde{h}=\min\{6\,\hat c_{\mathrm{obs}},\,10\,c_{\mathrm{arena}}\}$ & -- \\
\bottomrule
\end{tabular}

\caption{
Reward terms for all experiments.
}
\label{tab:rewards}\label{tab:walk_rewards}\label{tab:crawl_rewards}\label{tab:crazy_rewards}
\end{table}

\section{Training details}\label{sec:training-parameters}

Algorithm~\ref{alg:rsgf-ppo} proposes a practical implementation of RSGF-RL as used in Section~\ref{sec:results}.
It summarizes how the estimates $\Vh$, $\gh1$ and $\gh0$ entering~\eqref{eq:R-hat-definition} are computed at each iteration.
At iteration $t$, the policy $\pi_{\btheta_t}$ is rolled out in $N$ parallel environments for $T_{\mathrm{r}}$ steps each (cf.~Table~\ref{tab:experiment-params}); environments that terminate are reset and keep running, so each environment produces a segment containing one or more, possibly partial, episodes.
The constraint estimates are built from the discounted cost-to-go of $R_1$ along these segments, using a learned cost critic $W_\psi$ as the baseline $b_{1,t}$ in~\eqref{eq:gradient-expression}.
For the objective, instead of forming a gradient explicitly, we run a standard PPO-clip update starting from $\btheta_t$ and define $\gh0(\btheta_t)$ as the resulting parameter displacement divided by $h_t$, so that $-h_t\gh0(\btheta_t)$ is exactly the PPO step.
Although this quantity is not an unbiased estimate of $\nabla V_0(\btheta_t)$ (as it accumulates $J$ Adam-preconditioned minibatch steps evaluated along the PPO iterates), it is uniformly bounded, and hence satisfies~\eqref{eq:bounded-bias}-\eqref{eq:boundedness-estimates} for some $\tilde\zeta_0$, and we have found works well in practice. 
In this way, problem~\eqref{eq:R-hat-definition} acts as a filter on the PPO step: if $\bxi=-\gh0(\btheta_t)$ is feasible for~\eqref{eq:R-hat-definition}, then $\hat\Rc(\btheta_t)=-\gh0(\btheta_t)$ and the update coincides with the PPO iterate; otherwise, the PPO step is replaced by $h_t\hat\Rc(\btheta_t)$.

\begin{algorithm}[!htbp]
\caption{One iteration of RSGF-RL as implemented in Section~\ref{sec:results}}
\label{alg:rsgf-ppo}
\begin{algorithmic}[1]
\REQUIRE Policy parameter $\btheta_t$; cost critic $W_\psi$; PPO value critic and Adam state; stepsize $h_t$; parameters $\alpha,\beta,\gamma$
\STATE \textbf{Rollout:} for $n\in 1:N$, $k\in 0:T_{\mathrm{r}}-1$, sample $a_k^n\sim\pi_{\btheta_t}(\cdot\,|\,s_k^n)$ and record $(s_k^n,a_k^n,r_k^n,c_k^n,d_k^n)$, where $r_k^n=R_0(s_k^n,a_k^n,s_{k+1}^n)$, $c_k^n=R_1(s_k^n,a_k^n,s_{k+1}^n)$, and $d_k^n\in\{0,1\}$ indicates termination of an episode at step $k$
\STATE \textbf{Cost-to-go:} set $G_{T_{\mathrm{r}}}^n=0$ and $G_k^n=c_k^n+\gamma(1-d_k^n)\,G_{k+1}^n$ for $k=T_{\mathrm{r}}-1,\dots,0$
\STATE \textbf{Constraint value:} $\Vh(\btheta_t)=\frac{1}{NT_{\mathrm{r}}}\sum_{n=1}^{N}\sum_{k=0}^{T_{\mathrm{r}}-1}G_k^n$
\STATE \textbf{Cost critic:} fit $W_\psi$ to the targets $\{G_k^n\}$ by minimizing the squared error with $E_{\mathrm{c}}$ epochs of minibatch Adam
\STATE \textbf{Constraint gradient:}\\ \hspace*{1em}$\gh1(\btheta_t)=\frac{1}{NT_{\mathrm{r}}}\sum_{n,k}\big(G_k^n-W_\psi(s_k^n)\big)\,\nabla_{\btheta}\log\pi_{\btheta_t}(a_k^n\,|\,s_k^n)$
\STATE \textbf{PPO advantages:} compute GAE advantages $\hat A_k^n$ from $\{r_k^n\}$ and the value critic; set $\btheta^{(0)}=\btheta_t$
\FOR{$j=0,\dots,J-1$ \hfill ($J=$ epochs $\times$ minibatches)}
  \STATE 
  Obtain $\theta^{(j+1)}$ through Adam update on $\theta^{(j)}$, using gradient of $L^{\text{clip}}$ computed with minibatch $\mathcal{B}_j$.
\ENDFOR
\STATE \textbf{Objective gradient:} $\gh0(\btheta_t)=-\big(\btheta^{(J)}-\btheta_t\big)/h_t$
\STATE \textbf{RSGF direction:} compute $\hat\Rc(\btheta_t)$ from~\eqref{eq:R-hat-definition} in closed form~\cite[Lemma 3]{PM-AM-JC:25-l4dc}, or set $\hat\Rc(\btheta_t)=-\gh1(\btheta_t)/\beta$ if~\eqref{eq:R-hat-definition} is infeasible
\IF{$\bxi=-\gh0(\btheta_t)$ is feasible for~\eqref{eq:R-hat-definition}}
  \STATE $\btheta_{t+1}=\btheta^{(J)}$ \hfill (constraint inactive: $\hat\Rc(\btheta_t)=-\gh0(\btheta_t)$)
\ELSE
  \STATE $\btheta_{t+1}=\btheta_t+\sigma h_t\hat\Rc(\btheta_t)$, where $\sigma=2^{-m}$ and $m$ is the smallest integer in $0:4$ such that $\overline{\mathrm{KL}}(\pi_{\btheta_t}\,\|\,\pi_{\btheta_{t+1}})\le\mathrm{KL}_{\max}$ ($m=4$ if none)
\ENDIF
\end{algorithmic}
\end{algorithm}

In Algorithm~\ref{alg:rsgf-ppo}, $L^{\mathrm{clip}}$ is as defined in~\citep[Equation 9]{JS-FW-PD-AR-OK:17} and
$\overline{\mathrm{KL}}$ is the KL divergence averaged over the rollout states.
Adam uses gradient-norm clipping, 
and its moment estimates are carried over across iterations.
The cost critic $W_\psi$ has the same architecture as the value critic 
and receives the same observations.

\begin{table}[!htbp]
\centering
\footnotesize
\setlength{\tabcolsep}{4pt}
\renewcommand{\arraystretch}{1.1}
\begin{tabular}{@{} l *{3}{>{\raggedright\arraybackslash}p{0.235\linewidth}} @{}}
\toprule
& \textbf{Squat./crawl.} & \textbf{Hum.\ walking} & \textbf{Drone flight} \\
\midrule
\multicolumn{4}{@{}l}{\textit{Training parameters}} \\
Actor/Critic MLP size & (512, 256, 128) & (256, 128, 64) & (256, 256, 256) \\
Activation function  & ELU & ELU & ELU \\
Entropy coef.        & 0.005 & 0.003 & 0.001 \\
Epochs / minibatches & 5 / 4 & 5 / 4 & 5 / 4 \\
$J$ & 20 & 20 & 20 \\
$E_c$ & 4 & 4 & 4 \\
$KL_{\text{max}}$ & 0.08 & 0.08 & 0.08 \\
PPO learning rate$^{*}$  & $10^{-3}$ & $3\times10^{-4}$ & $3\times10^{-4}$ \\
PPO clip             & 0.2 & 0.2 & 0.2 \\
Discount factor $\gamma$ & 0.99 & 0.99 & 0.99 \\
GAE parameter $\lambda$  & 0.95 & 0.95 & 0.95 \\
Rollout (steps/env)  & 24 & 32 & 24 \\
Parallel envs        & 4096 & 2048 & 8192 \\
Total environment steps & $\approx (5.9 / 7.9) \times 10^8$ & $\approx 5.2\times 10^8$ & $\approx 7.9 \times 10^8$ \\
Safety constraint    & Base height $\geq 0.4$~m / Base and head $\leq 0.2$~m
                     & Base height $\geq 0.67$~m, $|\text{roll}| \leq 0.35$~rad, $|\text{pitch}| \leq 0.50$~rad, CoP in support polygon
                     & Distance to obstacles and arena edge $\geq 0.1$ \text{m}
                      \\
\midrule
\multicolumn{4}{@{}l}{\textit{RSGF-RL parameters}} \\
$\{ h_t \}$  & constant, 0.5 & constant, 0.5 & constant, 0.15 \\
$\alpha$     & $2 \times 10^{-5}$ & $2 \times 10^{-5}$ & 0.1 \\
$\beta$      & 1 & 1 & 1 \\
\bottomrule
\end{tabular}
\caption{Training and RSGF-RL parameters for each experiment. All experiments use observation normalization.
$^{*}$Adaptive, except for drone flight. The reported values refer to the initial PPO learning rate.
}
\label{tab:experiment-params}

\bigskip
\begin{tabular}[t]{@{}lr@{}}
\toprule
\textbf{Parameter} & \textbf{Value} \\
\midrule
\multicolumn{2}{@{}l}{\textit{Returns and advantages}} \\
Reward discount $\gamma$ & 0.99 \\
Cost discount $\gamma_{\mathrm{c}}$ & 0.99 \\
Reward / cost GAE parameter $\lambda$ & 0.95 / 0.95 \\
Cost limit $d$ & 0 \\
\midrule
\multicolumn{2}{@{}l}{\textit{Critics}} \\
Critic learning rate & $10^{-3}$ \\
Critic epochs / minibatches & 5 / 4 \\
\bottomrule
\end{tabular}%
\hspace{2.5em}%
\begin{tabular}[t]{@{}lr@{}}
\toprule
\textbf{Parameter} & \textbf{Value} \\
\midrule
\multicolumn{2}{@{}l}{\textit{Trust-region policy step}} \\
KL trust-region limit $\delta$ & 0.01 \\
Conjugate-gradient iterations & 15 \\
Conjugate-gradient damping & 0.1 \\
Fisher-vector-product subsampling & 5 \\
Line-search attempts & 15 \\
Line-search decay & 0.8 \\
\bottomrule
\end{tabular}
\caption{CPO parameters for the squatting comparison in Figure~\ref{fig:squatting}.}
\label{tab:cpo-params}
\end{table}

Regarding the baseline comparison in Figure~\ref{fig:squatting}, we point out that PPO-Lagrangian follows the implementation in~\cite{openai2019safetystarteragents} (Lagrange multiplier $\lambda=\mathrm{softplus}(\vartheta)$ initialized to 1 and updated with Adam, learning rate 0.05), with all PPO settings matching the PPO baselines in Table~\ref{tab:experiment-params} and the cost limit applied to the mean per-step cost.
On the other hand, CPO~\citep{JA-DH-AT-PA:17} uses the parameters in Table~\ref{tab:cpo-params}.

\end{document}